\documentclass[lettersize,journal]{IEEEtran}
\usepackage{amsmath,amsfonts}
\usepackage{algorithmic}
\usepackage{algorithm}
\usepackage{array}
\usepackage{textcomp}
\usepackage{stfloats}
\usepackage{url}
\usepackage{verbatim}
\usepackage{graphicx}
\usepackage{cite}

\usepackage{bm}
\usepackage{amsmath} 
\usepackage{graphicx}
\usepackage{pdfpages}
\usepackage{booktabs}  %  引入三线表宏包
\usepackage{arydshln}
\usepackage{multirow}
\usepackage{subcaption}
\usepackage{makecell}

\begin{document}

\title{MicroDreamer: Efficient 3D Generation in $\sim$20 Seconds by Score-based Iterative Reconstruction}

% \author{IEEE Publication Technology,~\IEEEmembership{Staff,~IEEE,}

%         % <-this % stops a space
% \thanks{* Equal contribution.}% <-this % stops a space
% % \thanks{Manuscript received April 19, 2021; revised August 16, 2021.}
% }

\author{Luxi Chen$^*$,~Zhengyi Wang$^*$,~Zihan Zhou,~Tingting Gao,~Hang Su,~\IEEEmembership{Member,~IEEE,}~Jun Zhu,~\IEEEmembership{Fellow,~IEEE,}~Chongxuan Li$^\dagger$~\IEEEmembership{Member,~IEEE}

\thanks{This work was supported by Beijing Nova Program (No. 20230484416); NSF of China (No. 62076145); Beijing Natural Science Foundation (No. L247030); the Kuaishou Research Fund. The work was partially done at the Engineering Research Center of Next-Generation Intelligent Search and Recommendation, Ministry of Education.}
\thanks{$^*$ Equal contribution.}
\thanks{Luxi Chen, Zihan Zhou, and Chongxuan Li are with the Gaoling School of AI, Renmin University of China, and Beijing Key Laboratory of Big Data Management and Analysis Methods, Beijing 100872, China. E-mail:  clx1489@ruc.edu.cn; zhouzihan2@ruc.edu.cn; chongxuanli@ruc.edu.cn. $^\dagger$\emph{Corresponding author: Chongxuan Li.}}
\thanks{Zhengyi Wang, Hang Su, and Jun Zhu are with Dept. of Comp. Sci. \& Tech., BNRist Center, Tsinghua-Bosch Joint ML Center, Tsinghua University, Beijing 100084, China. E-mail: wang-zy21@mails.tsinghua.edu.cn; suhangss@tsinghua.edu.cn; dcszj@tsinghua.edu.cn}
\thanks{Tingting Gao is with Kuaishou Technology, Beijing, China. E-mail: lisize@kuaishou.com.}
}

% The paper headers
% \markboth{Chen et al.: MicroDreamer: Efficient 3D Generation in $\sim$20 Seconds by Score-based Iterative Reconstruction}%
% {Shell \MakeLowercase{\textit{et al.}}: A Sample Article Using IEEEtran.cls for IEEE Journals}

% \IEEEpubid{0000--0000/00\$00.00~\copyright~2021 IEEE}
% Remember, if you use this you must call \IEEEpubidadjcol in the second
% column for its text to clear the IEEEpubid mark.

\maketitle

\begin{abstract}
Optimization-based approaches, such as score distillation sampling (SDS), show promise in zero-shot 3D generation but suffer from low efficiency, primarily due to the high number of function evaluations (NFEs) required for each sample and the limitation of optimization confined to latent space. This paper introduces score-based iterative reconstruction (SIR), an efficient and general algorithm mimicking a differentiable 3D reconstruction process to reduce the NFEs and enable optimization in pixel space. Given a single set of images sampled from a multi-view score-based diffusion model, SIR repeatedly optimizes 3D parameters, unlike the single-step optimization in SDS. With other improvements in training, we present an efficient approach called MicroDreamer that generally applies to various 3D representations and 3D generation tasks. In particular, MicroDreamer is 5-20 times faster than SDS in generating neural radiance field while retaining a comparable performance and takes about 20 seconds to create meshes from 3D Gaussian splatting on a single A100 GPU, halving the time of the fastest optimization-based baseline DreamGaussian with significantly superior performance compared to the measurement standard deviation. Our code is available at \url{https://github.com/ML-GSAI/MicroDreamer}.
\end{abstract}


\section{Introduction}
\label{sec:intro}

% \todo{a clear outline of the method}
% \todo{the phrase “mimicking 3D reconstruction” somewhat confusing}

% message
% clear, precise
% evidence
% coherent
% writing
% review problems

\IEEEPARstart{R}{ecently}, optimization-based approaches~\cite{poole2022dreamfusion,wang2023score,lin2022magic3d,zhu2023hifa,chen2023fantasia3d,liu2023zero1to3,wang2024prolificdreamer,tang2023dreamgaussian,liang2023luciddreamer,sun2023dreamcraft3d,metzer2022latent,yi2023gaussiandreamer,liu2023sherpa3d,qiu2023richdreamer,wu2023reconfusion,tang2023stable,yu2023text,katzir2023noise,yang2023learn} particularly score distillation sampling (SDS)~\cite{poole2022dreamfusion,wang2023score} have emerged as promising avenues for 3D generation based on text-to-image diffusion models~\cite{sohl2015deep,ho2020denoising,song2021scorebased,rombach2022high,saharia2022photorealistic,ramesh2022hierarchical}.  
These approaches are appealing due to their minimal, or even zero reliance on 3D data, in contrast to the data-intensive requirements of feed-forward approaches~\cite{cao2023large,chen2023single,chen2023primdiffusion,jun2023shap,nichol2022point,liu2023meshdiffusion,hong2023lrm,muller2023diffrf,wang2023rodin,yariv2023mosaic,zhao2023michelangelo,zou2023triplane,li2023instant3d,liu2023one,liu2023one2345++,wang2023pf,wang2024crm,tang2024lgm,hong20243dtopia,xu2024instantmesh,wang2023slice3d,lan2024ln3diff,TripoSR2024,xu2024grm,wei2024meshlrm,wu2024unique3d,zhang2024gs,boss2024sf3d}. This advantage is particularly significant given that 3D data are costly and scarce. Despite their promising capabilities, optimization-based approaches suffer from low efficiency due to the extensive number of function evaluations (NFEs), i.e. forward passes of the diffusion model, required for each 3D object generation. Moreover, when adopting the latent diffusion model (LDM)~\cite{rombach2022high} framework,  most of these approaches can only compute loss in latent space rather than pixel space, which requires backpropagation through an encoder~\cite{kingma2014auto,van2017neural} and further lowers the efficiency. Even if the loss function can be written in a data-predicting form~\cite{zhu2023hifa}, this type of loss is challenging to optimize effectively (see evidence in Fig.~\ref{fig:sds-fail}). The fastest approach DreamGaussian~\cite{tang2023dreamgaussian} still requires about 40 seconds to generate a 3D object even employing 3D Gaussian splatting~\cite{kerbl20233d}. 

% \todo{Besides, in latent space require bp through encoder. It is hard in pixel one step prediction, pixel space reconstruction bad, hard to for uptaing.}

\begin{figure}
    \centering
    \includegraphics[width=1.0\linewidth]{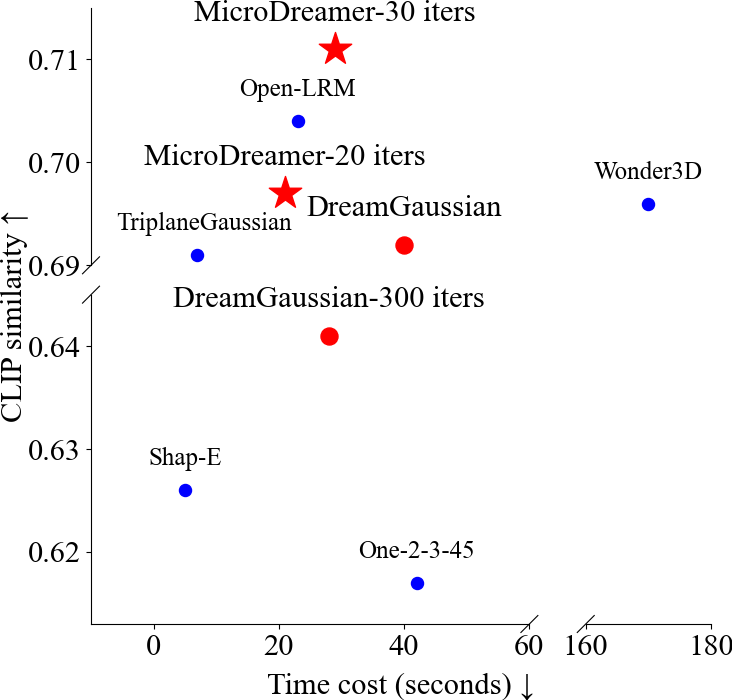}
    \caption{\textbf{MicroDreamer} surpasses the fastest optimization-based baseline DreamGaussian~\cite{tang2023dreamgaussian} in terms of both efficiency and sample quality. The optimization-based methods are highlighted in red. See Tab.~\ref{tab:quan-comp} for a comprehensive comparison with more baselines.}
      \label{fig:clip-time}
\end{figure}

\begin{figure*}[t]
  \centering
  \includegraphics[width=\linewidth]{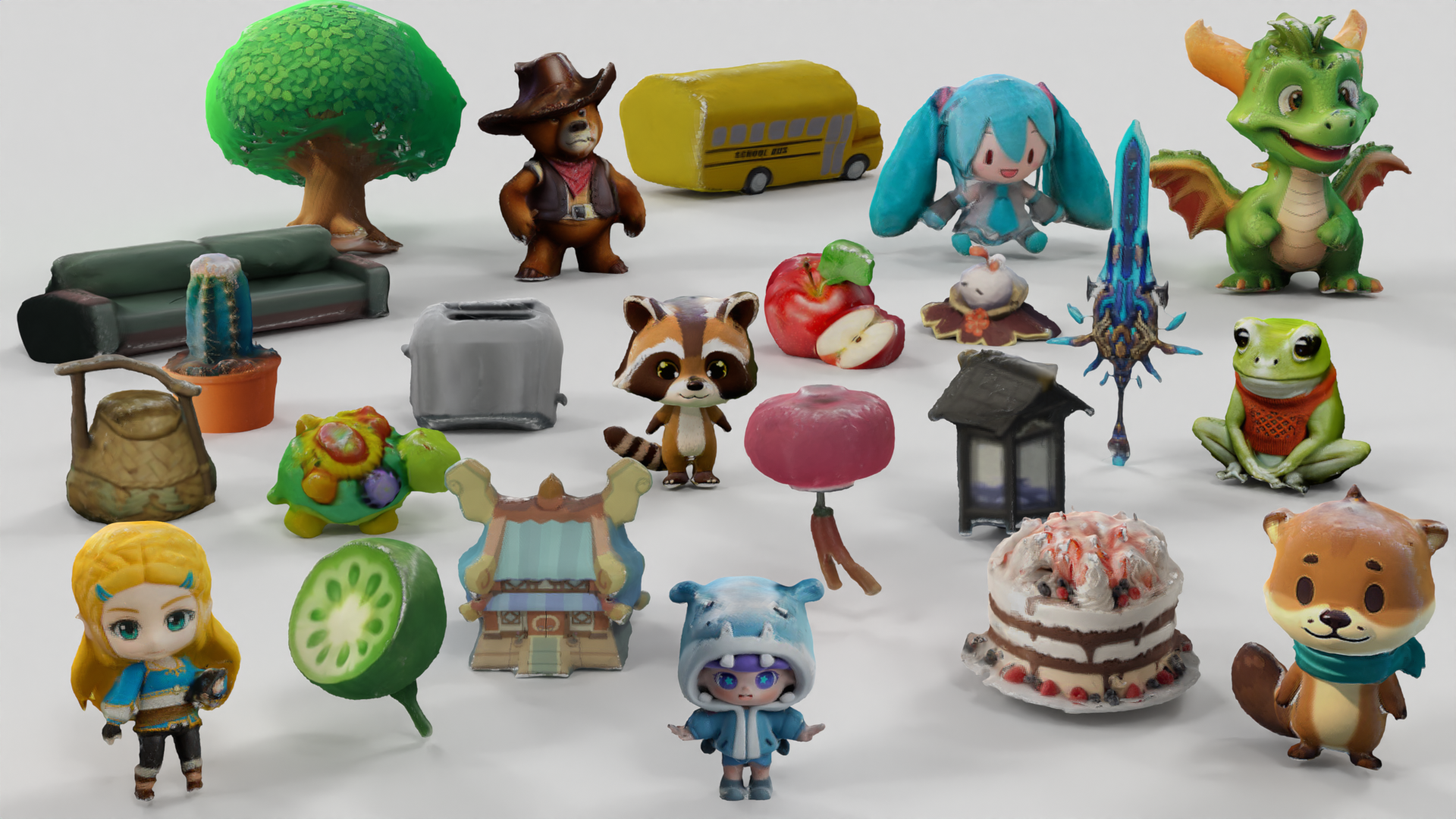}
  \caption{\textbf{MicroDreamer} can generate a high-quality mesh (as illustrated above) in about {\textbf{20 seconds}} on a single A100, built on a multi-view diffusion model {\textbf{without additional 3D data}}. See supplementary materials for 3D visualization.}
  \label{fig:intro}
  % \vspace{-.5cm}
\end{figure*}

% \todo{taking xxx time even bases 3dgs. The algorithm leads to xxx time in sota.}

In comparison, the multi-step reconstruction process of 3D representations that enable differentiable rendering, such as neural radiance field (NeRF)~\cite{mildenhall2021nerf,muller2022instant} and 3D Gaussian splatting (3DGS)~\cite{kerbl20233d}, produces 3D contents extremely fast because they do not involve large generative neural networks.
However, such approaches rely on true 3D data, i.e. abundant real multi-view images, making them unfeasible for text-to-3D and image-to-3D generation tasks. There exist works~\cite{liu2023syncdreamer,long2023wonder3d,shi2023zero123plus,lu2023direct2,li2024era3d,melas20243d,chen2024v3d,voleti2024sv3d} for 3D generation attempt to generate multi-views first to reconstruct 3D object directly. Still, such methods may require the diffusion model to simultaneously generate multi-view images with their corresponding 3D priors~\cite{long2023wonder3d,li2024era3d}, and they require longer than one minute to reconstruct a 3D object. (see Sec.~\ref{sec:related} for a review and Sec.~\ref{sec:experiment} for comparison). 
% or necessitate the generation of numerous but potentially inconsistent views from video diffusion~\cite{melas20243d,chen2024v3d}. 

To speed up the generation process, this paper presents an efficient and general 3D generation algorithm termed \textbf{score-based iterative reconstruction (SIR)}, leveraging reconstruction to reduce NFEs and enable optimization in pixel space. Like SDS, SIR iteratively updates 3D parameters, leveraging a multi-view diffusion model without additional 3D data or 3D prior. However, in each iteration, SIR distinguishes itself by repeatedly optimizing 3D parameters given a set of images produced by the diffusion model, mimicking the efficient 3D reconstruction process to reduce the total NFEs (see Fig.~\ref{fig:compare-sds-nerf}). To obtain 3D-consistent and high-quality images as the ground truth for better reconstruction in each iteration, we carefully design a hybrid forward process and a sampling process to refine the images rendered from the 3D object optimized in the latest iteration. Besides, even mapped back to pixel space through the decoder in LDM, the refined images are still of high quality for reconstruction, enabling optimization in pixel space to speed up further (see Fig.~\ref{fig:abla-optim-space}).

% \todo{ablation for detailed analysis. general configuration microdreamer.}
% We delve into critical components beyond the algorithm to ensure these components are compatible with SIR. 

We provide a general and compatible configuration for SIR, and the comprehensive system is named \textbf{MicroDreamer}, highlighting its efficiency for 3D generation. As detailed in Sec.~\ref{sec:microdreamer}, we introduce an initialization strategy for 3D objects, an annealed time schedule for diffusion, additional losses on reference images for image-to-3D, and a refinement procedure for deriving high-quality meshes from 3DGS.

% by conducting a detailed ablation analysis

% \todo{Comprehensive studies to show generality and efficiency. In particular, x}

Comprehensive studies demonstrate the generality and efficiency of our proposed method. In particular, SIR and MicroDreamer broadly apply to NeRF and 3DGS and both text-to-3D and image-to-3D tasks, as detailed in Sec.~\ref{sec:experiment}. Employing three widely adopted multi-view diffusion models~\cite{shi2023mvdream,stable123,wang2023imagedream}, we systematically compare SIR and SDS for NeRF generation. Retaining a comparable performance, SIR can accelerate the generation process by \textbf{5 to 20 times}. Besides, MicroDreamer can efficiently produce 3DGS and further refine it into high-quality meshes, delivering consistent 3D meshes in about \textbf{20 seconds} on a single A100 GPU -- about twice as fast as the most competitive optimization-based alternatives, DreamGaussian~\cite{tang2023dreamgaussian}, with significantly (compared to the measurement standard deviation) superior performance. Remarkably, MicroDreamer is on par with speed compared to representative feed-forward methods~\cite{openlrm} trained on extensive 3D data, with a very competitive CLIP similarity~\cite{radford2021learning}.

\begin{figure*}[tb]
    \centering
    \includegraphics[width=\linewidth]{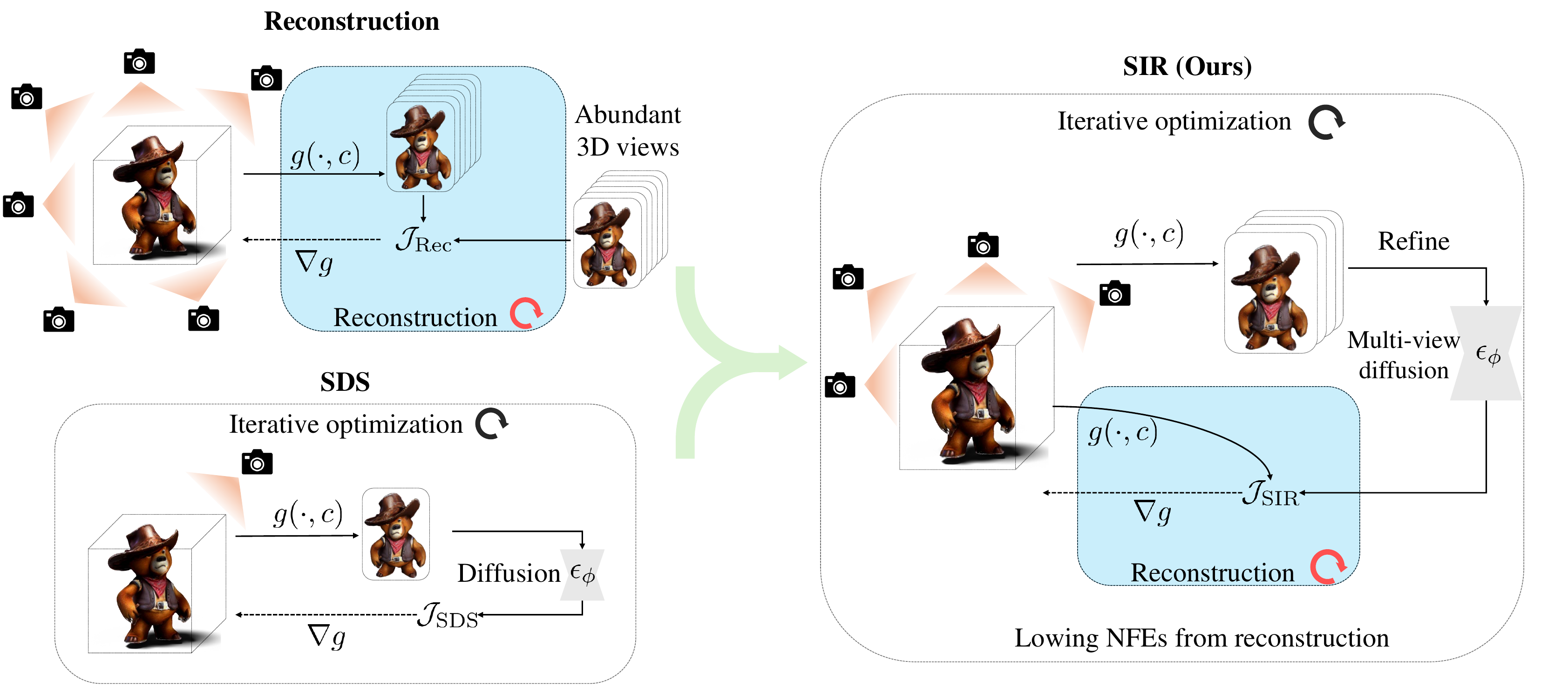}
    \caption{\textbf{Overview of SIR.} 
    SIR is an optimization-based 3D generation method that marries the strengths of reconstruction and iterative optimization. SIR reutilizes the samples from diffusion multiple times through reconstruction, reducing the total NFEs, enabling optimization in pixel space, and improving efficiency.
    }
    \label{fig:comp-plot}
\end{figure*}

% $\mathcal{E}$, $\mathcal{D}$ and $\epsilon_{\phi}$ are the encoder, decoder, and diffusion model in  LDM~\cite{rombach2022high}.
%     The solid and dashed arrows represent forward 
%  and backward passes respectively. The computation bottlenecks are highlighted in red. SIR is efficient because it lowers the NFEs in total and enables optimization in the pixel space.

\section{Related work}
\label{sec:related}
\emph{Optimization-based 3D generation.}
Built upon text-to-image diffusion models, optimization-based approaches~\cite{poole2022dreamfusion,wang2023score,lin2022magic3d,zhu2023hifa,chen2023fantasia3d,liu2023zero1to3,wang2024prolificdreamer,tang2023dreamgaussian,liang2023luciddreamer,sun2023dreamcraft3d,metzer2022latent,yi2023gaussiandreamer,liu2023sherpa3d,qiu2023richdreamer,wu2023reconfusion,tang2023stable,yu2023text,katzir2023noise,yang2023learn} usually generate 3D objects without additional 3D data. Among them, the most relevant line of work~\cite{poole2022dreamfusion,wang2024prolificdreamer,liang2023luciddreamer,zhu2023hifa,wang2023score,yu2023text,katzir2023noise,yang2023learn} proposes various distillation algorithms. Besides SDS~\cite{poole2022dreamfusion}, ProlificDreamer~\cite{wang2024prolificdreamer} proposes variational score distillation (VSD) to produce high-fidelity 3D objects via variational inference. LucidDreamer~\cite{liang2023luciddreamer} incorporates DDIM inversion to strengthen the forward process. ReconFusion~\cite{wu2023reconfusion} utilizes diffusion priors to generate novel views for single-step optimization in sparse view reconstruction tasks, aiming to enhance the quality of scene reconstruction. Though promising, many of these methods suffer from inefficiency problems due to large NFEs and optimization in latent space. Distinct from existing algorithms, SIR mimics 3D reconstruction processes to optimize the 3D parameters multiple times given a set of images produced by diffusion. Our experiments demonstrate that SIR is 5-20 times faster than SDS~\cite{poole2022dreamfusion} on NeRF and twice as fast as the most competitive baseline DreamGaussian~\cite{tang2023dreamgaussian} with better generation quality on meshes.

\emph{Feed-forward 3D generation.}
Contrary to optimization-based methods, feed-forward methods~\cite{cao2023large,chen2023single,chen2023primdiffusion,jun2023shap,nichol2022point,liu2023meshdiffusion,hong2023lrm,muller2023diffrf,wang2023rodin,yariv2023mosaic,zhao2023michelangelo,zou2023triplane,li2023instant3d,liu2023one,liu2023one2345++,wang2023pf,wang2024crm,hong20243dtopia,xu2024instantmesh,wang2023slice3d,lan2024ln3diff,TripoSR2024,xu2024grm,wei2024meshlrm,wu2024unique3d,zhang2024gs,boss2024sf3d} use large-scale 3D datasets~\cite{deitke2023objaverse,deitke2023objaversexl} training to achieve a process that can directly generate 3D objects. Early methods are characterized by their speed but often produce lower-quality 3D structures with simple textures. Several recent methods~\cite{openlrm,hong2023lrm,zhao2023michelangelo,zou2023triplane,TripoSR2024,boss2024sf3d} have exhibited the feasibility of training a Transformer utilizing more data to achieve reliable 3D content from a single image. Generating 3D content from a single view can be quite challenging. Therefore, several follow-up works~\cite{li2023instant3d,liu2023one,liu2023one2345++,xu2024grm,wu2024unique3d,wang2024crm,tang2024lgm,xu2024instantmesh} have leveraged multi-view diffusion models to generate multiple views as input and then trained a feed-forward model to produce 3D content from these views. Remarkably, MicroDreamer is on par with speed compared to such methods trained on extensive 3D data, with a very competitive 3D quality measured by CLIP similarity~\cite{radford2021learning}.

\emph{Multi-view prediction based 3D generation.}
There is also a line of work~\cite{liu2023syncdreamer,long2023wonder3d,shi2023zero123plus,lu2023direct2,li2024era3d,melas20243d,chen2024v3d,voleti2024sv3d} dedicated to enhancing the output of multi-view diffusion models by training on 3D datasets to reconstruct 3D objects using a single reconstruction process with no or few iterations. Some approaches like Wonder3D~\cite{long2023wonder3d} typically rely on 3D prior knowledge and are limited to specific 3D representations with a long reconstruction time. Another aspect of works like IM-3D~\cite{melas20243d} and V3D~\cite{chen2024v3d} involves fine-tuning the video diffusion model using 3D data and employing the generated 3D-aware multi-view images as ground truth for reconstruction. The efficiency of these methods is limited by the sampling efficiency of the video diffusion model and the long reconstruction time. Consequently, the total time required to generate a 3D object using these methods typically exceeds one minute. In contrast, thanks to the carefully designed iterative process in SIR, MicroDreamer is more efficient and applies to various 3D representations (see Tab.~\ref{tab:quan-comp} for comparison).

\section{Background}
\label{sec:background}

We present background on 3D representations, diffusion models, multi-view diffusion models, and current optimization-based 3D generation methods sequentially.

\subsection{3D representation}
\label{sec:3D-repre}
Neural radiance fields~\cite{mildenhall2021nerf,muller2022instant} (NeRF) and 3D Gaussian splatting~\cite{kerbl20233d} (3DGS) have emerged as popular 3D representations. NeRF employs an MLP to predict the colors and density of the input space coordinates. 3DGS consists of multiple 3D Gaussians parameterized by the colors, centers, scales, and rotation quaternions. We denote the corresponding tunable parameters in both representations as $\theta$. Given camera poses $c$, both approaches define a differentiable rendering process, denoted by $g(\theta, c)$. They are proven efficient and effective in 3D reconstruction~\cite{mildenhall2021nerf,kerbl20233d} and generation~\cite{poole2022dreamfusion,tang2023dreamgaussian}.

\subsection{Diffusion models}

A diffusion model~\cite{sohl2015deep,ho2020denoising,song2021scorebased} consists of a forward process and a sampling process. 
The forward process gradually adds Gaussian noise to an input image from time $0$ to $T$. For any $t\in (0, T)$, the noise-adding process can be written as follows:
\begin{align}
    x_t  &= \alpha_t x_0 + \sigma_t \epsilon \nonumber \\
    &:=\textrm{Noise-adding}(x_0,0\rightarrow t), \, \epsilon \in \mathcal{N}(0,I),
   \label{eq:noise-adding}
\end{align}
where the coefficients $\alpha_t$ and $\sigma_t$ form the noise schedule of the diffusion model. A noise prediction network $\epsilon_{\phi}(x_t, t)$ with parameters $\phi$ is trained to predict the noise in the input $x_t$ with the corresponding noise level, i.e. time $t$.
Plugging in the noise prediction network into Eq.~(\ref{eq:noise-adding}), we can solve $x_0$ from a noisy image $x_t$ of time $t$ by a single-step prediction as follows:
\begin{align}
    \hat{x}^t_{0} =   \frac{1}{{\alpha}_t} x_t - {\frac{\sigma_t}{{\alpha}_t}} \epsilon_{\phi} (x_t, t), \label{eq:x0t}
\end{align}
which is an efficient approximation of the original input. 

Instead of using Eq.~(\ref{eq:x0t}) directly, the sampling process of diffusion gradually denoises through the noise prediction network and generates images from pure Gaussian noise. Among existing samplers~\cite{liu2022pseudo,bao2022analytic,bao2022estimating,lu2022dpm,lu2023dpmsolver,zhao2023unipc}, the denoising diffusion implicit models (DDIM)~\cite{song2020denoising} facilitate a sequence of samplers with random noise control $\eta$. When $\eta=0$, it solves the equivalent probability flow ordinary differential equation (ODE)~\cite{song2021scorebased} of the diffusion model and enjoys a fast sampling process with a small number of function evaluations (NFEs)\footnote{Throughout the paper, we refer to the number of forward passes through $\epsilon_{\phi}$ as NFEs.}. In this setting, the one-step sampling of DDIM is given by:
 % (see Sec.~\ref{sec:related} for a review)
% \begin{align}
%     x_{t-1} &= \alpha_{t-1}\left(\frac{x_t-\sigma_t\epsilon_{\phi} (x_t, t)}{\alpha_t}\right)+\sqrt{\sigma_{t-1}^2-\eta^2\Tilde{\beta}_t^2 }\epsilon_{\phi} (x_t, t)+\eta\Tilde{\beta}_t \epsilon \label{eq:sampling}\nonumber\\
%     &:={\textrm{Sampler}}(x_{t},t\rightarrow t-1).\\
%     \epsilon &\sim \mathcal{N}(0,I),\quad \beta_t=\frac{\sigma_{t-1}}{\sigma_t}\sqrt{1-\frac{\alpha_t^2}{\alpha^2_{t-1}}}\nonumber
% \end{align}
\begin{align}
    x_{t-1}&=\frac{\alpha_{t-1}}{\alpha_t}x_{t}+\left(\sigma_{t-1}-\frac{\alpha_{t-1}}{\alpha_{t}}\sigma_t\right) \epsilon_{\phi} (x_t, t) \nonumber\\
    &:={\textrm{Sampler}}(x_{t},t\rightarrow t-1).\label{eq:sampling}
\end{align}
We refer the readers to the original paper~\cite{song2020denoising} for the formula of $\eta > 0$. For $\eta=1$, it represents a standard SDE sampler~\cite{ho2020denoising}, with a higher tolerance for mismatches in the distributions of latent variables~\cite{lu2022maximum,nie2023blessing}.

Besides, when $\eta=0$ there is an inverse process (called DDIM inversion) that maps the distribution of images to the same distributions of noisy images as in Eq.~(\ref{eq:noise-adding}) but it can maintain the unique feature of an input image and reconstruct it accurately through the corresponding DDIM sampler in Eq.~(\ref{eq:sampling}) (see Fig.~\ref{fig:refine}a). The (one-step) DDIM inversion~\cite{song2020denoising} is formulated as follows:
\begin{align}
    x_{t+1}&=\frac{\alpha_{t+1}}{\alpha_t}x_{t}+\left(\sigma_{t+1}-\frac{\alpha_{t+1}}{\alpha_{t}}\sigma_t\right) \epsilon_{\phi} (x_t, t)  \nonumber\\
    &:={\textrm{Inversion}}(x_{t},t\rightarrow t+1).\label{eq:inversion}
\end{align}

\subsection{Multi-view diffusion models}
 
After being trained on a modest amount of 3D data, diffusion models can generate 3D-consistent multi-view, known as multi-view diffusion models. Among them, MVDream~\cite{shi2023mvdream} takes text inputs and outputs multi-view images consistent in 3D. In contrast, Zero-1-to-3~\cite{liu2023zero1to3} and ImageDream~\cite{wang2023imagedream} focus on the Image-to-3D task, taking an additional reference image as input. These models output new viewpoint images consistent with the reference image. This paper directly utilizes these pre-trained multi-view diffusion models without any further fine-tuning. Notably, such multi-view diffusion models cannot provide sufficient consistent multi-view images for 3D reconstruction directly (see Fig.~\ref{fig:abla0}). Given that the multi-view diffusion models in~\cite{liu2023syncdreamer,long2023wonder3d,shi2023zero123plus,lu2023direct2,melas20243d,li2024era3d} often incorporate additional 3D priors and do not align with our setting in Sec.~\ref{sec:microdreamer}, we choose not to employ them in this paper.

% For more details, please refer to the original papers and the source code in our supplementary materials.

\subsection{Optimization-based algorithms for 3D generation}

Built upon (multi-view) diffusion models, optimization-based methods (see Sec.~\ref{sec:related} for a review) aim to generate 3D content in a zero-shot manner. Among them, 
score distillation sampling (SDS)~\cite{poole2022dreamfusion,wang2023score} is the most representative and popular approach. Formally, 
denoting the rendered images as $x = g(\theta,c)$, 
SDS repeats adding noise to $x$ according to Eq.~(\ref{eq:noise-adding}) and updating the 3D parameters $\theta$ by 
\begin{align}
 &\nabla_{\theta} \mathcal{J}_{\textrm{SDS}}(x=g(\theta); \phi) \nonumber\\
 := & \mathbb{E}_{t} \left [ w(t)  ( \epsilon_{\phi} (\alpha_t x + \sigma_t \epsilon, t) - \epsilon) \frac{\partial x}{\partial \theta} \right],\label{eq:SDS}
\end{align}
where $w(t)$ is a fixed weighting function, we omit the dependency on the prompt $y$ and camera $c$ for simplicity.
Notably, by reparameterization~\cite{zhu2023hifa} according to Eq.~(\ref{eq:x0t}), SDS has an equivalent data-prediction form:
\begin{align}
    \nabla_{\theta} \mathcal{J}_{\textrm{SDS}}(x=g(\theta); \phi) = \mathbb{E}_{t} \left [ \frac{w(t)\alpha_t}{\sigma_t}  ( x-\hat{x}_0^t) \frac{\partial x}{\partial \theta} \right].\label{eq:SDS-x0}
    % \mathcal{J}_{\textrm{SDS}}(x=g(\theta); \phi) = \mathbb{E}_{t} \left [ \frac{w(t)\alpha_t}{\sigma_t}  \| x-\hat{x}_0^t\|_2^2\right].
\end{align}

\section{Score-based iterative reconstruction}
\label{sec:method}

% SIR combines both the advantages of traditional 3D reconstruction and optimization methods.
% inspired by classical 3D reconstruction

% We first analyze factors that efficiency bottleneck of existing work in Sec.~\ref{sec:sds-bottleneck}

We first analyze factors contributing to the efficiency bottleneck of existing work in Sec.~\ref{sec:sds-bottleneck},  motivating score-based iterative reconstruction (SIR), an efficient and versatile algorithm combining the advantages of differentiable 3D reconstruction and optimization methods in Sec.~\ref{sec:sir}. We introduce a carefully designed diffusion-based process to produce refined multi-view images as ground truth for better reconstruction in Sec.~\ref{sec:refine}. Besides, we demonstrate how SIR facilitates the optimization of 3D content directly within pixel space in Sec.~\ref{sec:sir-pixel-space}.

% SIR capitalizes on the efficiency of 3D reconstruction and aims to minimize the number of function evaluations (NFEs) typically required by existing methods while ensuring comparable quality through iterations.

\subsection{Efficiency bottleneck of existing work}
\label{sec:sds-bottleneck}

\begin{figure}
    \centering
    \includegraphics[width=1.0\linewidth]{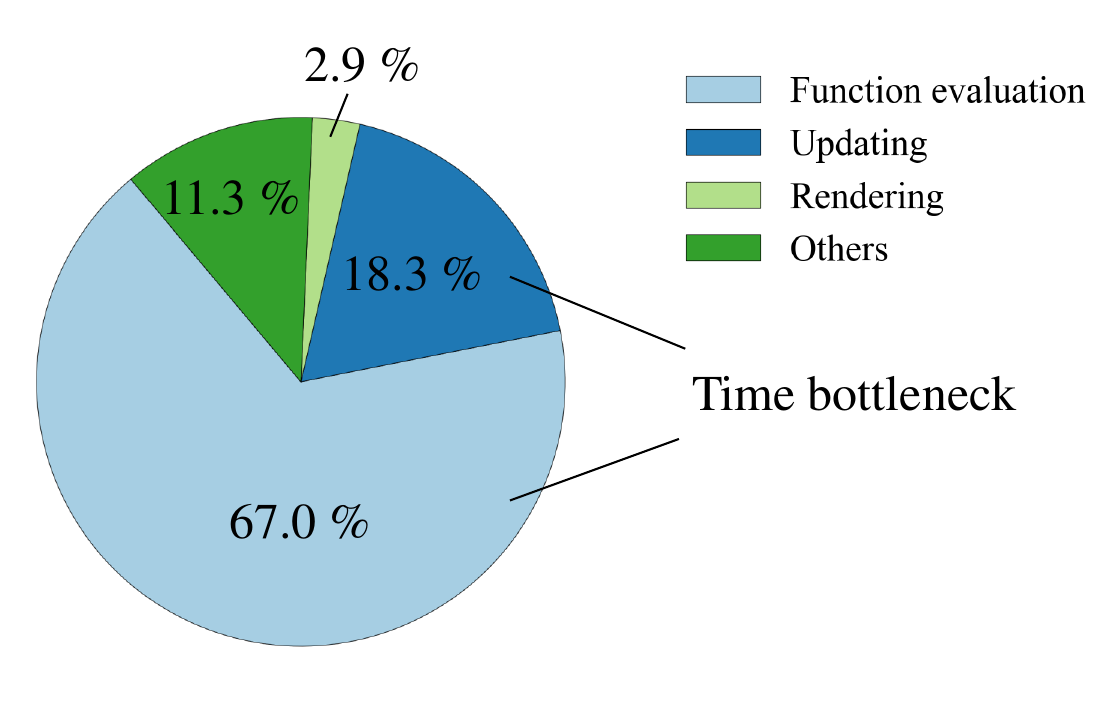}
    \caption{\textbf{Time proportion in SDS optimization.} We record the time proportions of all components in SDS on 3DGS. The bottleneck lies in the large NFEs and updating in latent space.}
    \label{fig:sds-bottleneck}
\end{figure}

As a motivation of SIR, we analyze the efficiency bottleneck of existing optimization-based methods~\cite{poole2022dreamfusion,wang2023score,zhu2023hifa,liang2023luciddreamer,yu2023text,katzir2023noise,wang2024prolificdreamer}. We take the widely adopted SDS~\cite{poole2022dreamfusion} as a representative example and our analysis also applies to other algorithms.

On the one hand, according to Eq.~(\ref{eq:SDS}), SDS iteratively optimizes the 3D parameters based on a 2D diffusion. It necessitates high NFEs because it requires a forward pass of the 2D diffusion in each update of the 3D parameters. On the other hand, when employed within the latent diffusion model (LDM)~\cite{rombach2022high} upon a variational auto-encoder (VAE)~\cite{kingma2014adam,van2017neural}, SDS computes the loss in latent space. This process requires backpropagation through the VAE encoder, further lowering generation efficiency. 
These arguments are validated by the empirical results in Fig.~\ref{fig:sds-bottleneck} quantitatively.

As for the latent space optimization, we emphasize that although SDS has a corresponding data-prediction form in Eq.~(\ref{eq:SDS-x0}), it is nontrivial to map the predicted $x_0^{(t)}$ back to pixel space through the VAE decoder for loss calculation. This is because the single-step prediction in Eq.~(\ref{eq:x0t}) yields poor samples in the pixel space, leading to suboptimal 3D results (See experiments in Tab.~\ref{fig:sds-fail} of Sec.~\ref{sec:experiment}).

% sampling time and portion. 
% 扇形图，大小表示采样时间，占比；结论是要想快怎么办；对比我们的：为啥快了，NFE 和 pixel。

\subsection{Mimicking 3D reconstruction to reduce the NFEs}
\label{sec:sir}

Motivated by the analysis of SDS, we propose SIR to reduce NFEs and enable optimization in pixel space to enhance efficiency. Naturally, reutilizing the outcomes of diffusion for successive updates of the 3D object—mimicking the process of differentiable 3D reconstruction could substantially decrease the overall NFEs required and enable optimization in pixel space. However, 3D reconstruction typically requires a sufficient number of consistently aligned multi-view images, which cannot be directly obtained from the current multi-view diffusion models~\cite{liu2023zero1to3,shi2023mvdream,wang2023imagedream,stable123} (see Sec.~\ref{sec:related} for a review and discussion). Therefore, similar to existing optimization-based methods, we introduce a reconstruction-based algorithm with multiple iterations of optimizing dubbed \textbf{score-based iterative reconstruction (SIR)}, detailed as follows.

Formally, SIR consists of $K$ reconstruction iterations. 
In the $k$-th iteration, where $k=0,1,\ldots, K-1$, given initial 3D parameters $\theta_{0}^{(k)}$, we randomly select several camera poses $c^{(k)}$ (detailed in Sec.~\ref{sec:microdreamer}) and employ rendering function  $g(\cdot, \cdot)$ to obtain multi-view images of the current 3D object as follows:
\begin{align}
    x^{(k)}=g(\theta^{(k)}_0,c^{(k)}). \label{eq:rendering}
\end{align}
For simplicity, let $x^{(k)}$ be a vector by flattening all images and concatenating them together, and so do any subsequent set of multi-view images.

As detailed in Sec.~\ref{sec:refine}, 
$x^{(k)}$ is then refined via a pre-trained multi-view diffusion model~\cite{shi2023mvdream,liu2023zero1to3,wang2023imagedream,stable123}. The outcome, denoted as 
$\hat{x}^{(k)}$, serves as the ground truth for reconstruction in the current iteration.
In particular, starting from $\theta_0^{(k)}$, keeping the camera poses $c^{(k)}$ unchanged, we optimize the 3D parameters w.r.t. the following reconstruction loss for $I$ steps:
\begin{align}
     \mathcal{J}_{\textrm{SIR}}(\theta; c^{(k)}, \hat{x}^{(k)}) = \| g(\theta, c^{(k)})-\hat{x}^{(k)}\|, \label{eq:sir}
\end{align}
where $\|\cdot\|$ can be any proper norm operator in principle while we choose $\ell_1$ norm in our experiments. We use the final 3D parameters of the $k$-th iteration as the initialization of the next one, i.e.
$\theta^{(k+1)}_0=\theta^{(k)}_{I}$ and outputs $\theta^{(K)}_0$ finally. See Sec.~\ref{sec:microdreamer} for the initialization of the $0$-th iteration, i.e. $\theta_0^{(0)}$.

Here we assume the rendered images and those sampled by the diffusion model have the same dimensional for simplicity. In Sec.~\ref{sec:sir-pixel-space} we will further discuss how to deal with the latent diffusion model (LDM)~\cite{rombach2022high}. Before that, we discuss how to obtain 3D-consistent and high quality $\hat{x}^{(k)}$ from $x^{(k)}$ for reconstruction in Sec.~\ref{sec:refine}.

\begin{figure}[tb]
    \centering
    \includegraphics[width=1.0\linewidth]{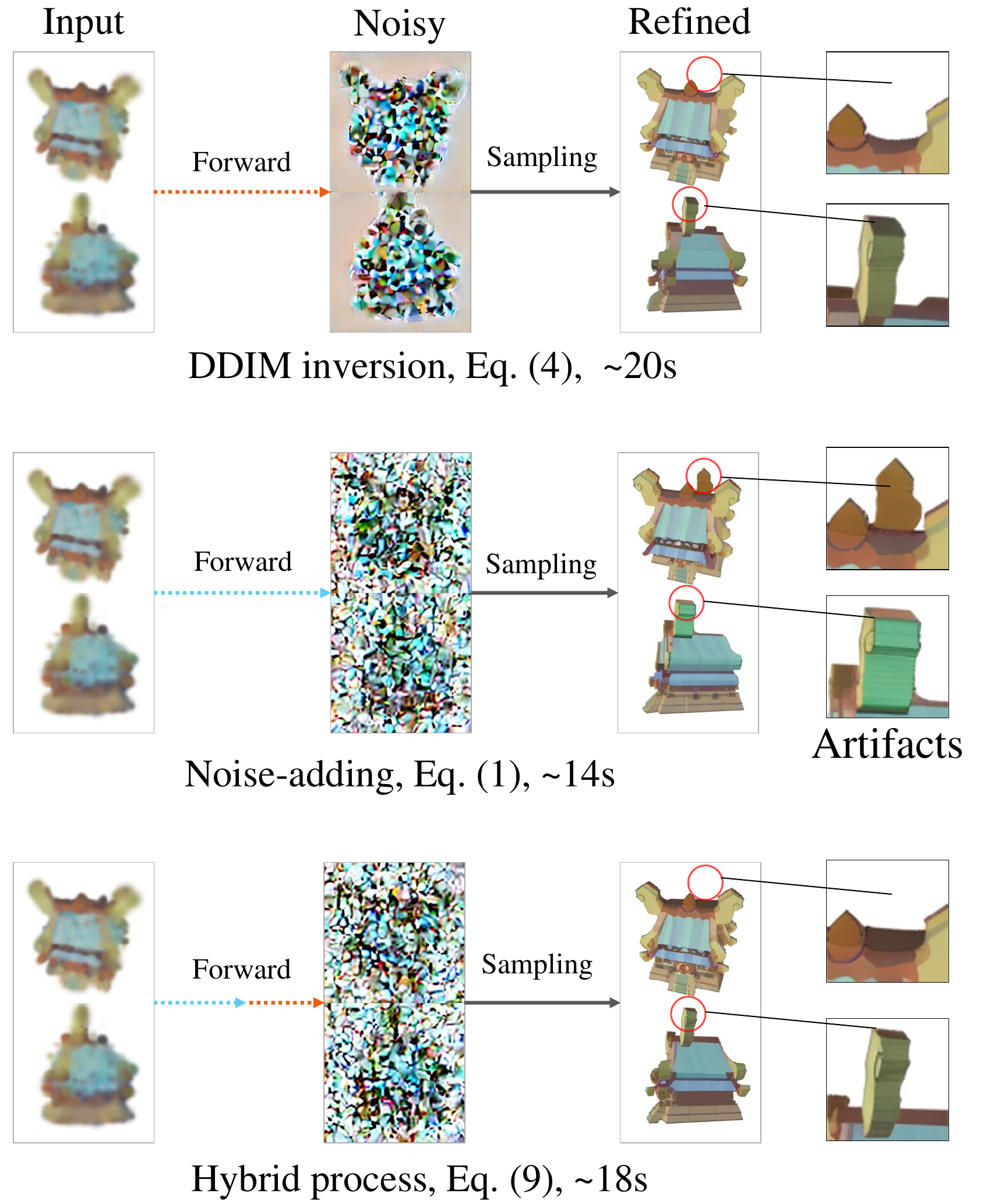}
    \caption{The \textbf{hybrid forward process} is more efficient than DDIM inversion and generates better samples than noise-adding. We present the final results and sampling time on 20 iterations of SIR for three forward processes. Notably, the Noise-adding process may generate artifacts that contain unexpected elements compared to the input.}
    \label{fig:refine}
\end{figure}

\subsection{Refining multi-view images for reconstruction by diffusion}
\label{sec:refine}

\begin{algorithm}[t] 
    \caption{Score-based iterative reconstruction (SIR)}
    \label{alg:sir} 
	\begin{algorithmic}[1]
	\STATE \textbf{Input: }{The number of iterations $K$, an initial 3D object $\theta_0^{(0)}$, the number of reconstruction steps $I$ and a set of camera poses $\{c^{(k)}\}_{k=0}^{K-1}$.} 
	\STATE \textbf{Output: }{A final 3D content $\theta^{(K)}_{0}$.}
        % \Output{A final 3D content $\theta_K$}
        % Initialize 3D object $O$ \;
	 \FOR {$k$ from $0$ to $K-1$}
        \STATE Render $N$ images $x^{(k)}=g(\theta^{(k)}_{0},c^{(k)})$
        \STATE Obtain noisy $\Tilde{x}^{(k)}$ from forward process for $x^{(k)}$
        \STATE Obtain refined $\hat{x}^{(k)}$ from sampling process for $\Tilde{x}^{(k)}$
        \FOR{$i$ from $0$ to $I-1$}
        \STATE  Compute loss $L=\| g(\theta,c^{(k)})-\hat{x}^{(k)}\|$
        \STATE  Compute the gradient $\nabla L$, update $\theta^{(k)}_{i}$ to $\theta^{(k)}_{i+1}$
        \ENDFOR
        \ENDFOR
  \end{algorithmic}
\end{algorithm}

We treat $x^{(k)}$ as noise-free (i.e., at time $0$ of the diffusion process) but low-quality images, refined by a forward process followed by a sampling process. 

The diffusion model has two theoretically equivalent forward processes~\cite{song2021scorebased}: the noise-adding process modeled by stochastic differential equations (SDEs) in Eq.~(\ref{eq:noise-adding}) and the inversion process based on the probability flow ordinary differential equations (ODEs) in Eq.~(\ref{eq:inversion}). In comparison, the noise-adding process is more efficient without function evaluation, but sampling after the noise-adding process may produce unexpected artifacts (see Fig.~\ref{fig:refine}). The inversion process can better preserve the 3D consistency and overall information of the current 3D object, but it necessitates more NFEs with low efficiency. We carefully design a hybrid forward process that initially adds noise and then performs DDIM inversion. Compared to common noise-adding and inversion processes, the hybrid forward process achieves a better balance in quality and efficiency, as Fig.~\ref{fig:refine} shows. 

% We mention that which achieves
% \todo{a better trade-off on quality and efficiency, compared to common noise-adding and inversion processes (see Fig.~xx).}

Specifically, the hybrid forward process adds noise to time $t_1^{(k)}$ first and then performs DDIM inversion~\cite{song2020denoising} to time $t_2^{(k)}$, where $t_1^{(k)} \in (0, T)$ and $t_2^{(k)} \in [t_1^{(k)}, T)$ are hyperparameters\footnote{Note that if $t^{(k)}_1=t^{(k)}_2$, the process is purely adding noise.}. In contrast to random sampled $t_2^{(k)}$ in existing algorithms including SDS~\cite{poole2022dreamfusion}, we adopt a linearly decreased schedule for $t_2^{(k)}$ as $k$ progresses, detailed in Sec.~{\ref{sec:microdreamer}}.
Formally, the process is defined as: 
\begin{align}
    \Tilde{x}^{(k)}=\textrm{Inversion}(\textrm{Noise-adding}(x^{(k)}, 0 \rightarrow t_1^{(k)}), t_1^{(k)} \rightarrow t_2^{(k)}), \label{eq:forward}
\end{align}
where $\textrm{Noise-adding}(\cdot)$ is defined in Eq.~(\ref{eq:noise-adding}) and $\textrm{Inversion}(\cdot)$ is  defined in Eq.~(\ref{eq:inversion}). The ablation in Fig.~\ref{fig:abla3} shows the effectiveness of our hybrid forward process.

Note that generally $x^{(k)}$ does not follow the model distribution defined by the diffusion and the resulting $\Tilde{x}^{(k)}$ does not strictly adhere to the marginal distribution at the corresponding time. Nevertheless, we still use existing sampling algorithms to obtain refined images $\hat{x}^{(k)}$ from $\Tilde{x}^{(k)}$ as follows:
\begin{align}
    \hat{x}^{(k)} ={\textrm{Sampler}}(\Tilde{x}^{(k)}, t_2^{(k)} \rightarrow 0). \label{eq:sir-sampling}
\end{align}
Although other advanced sampling methods~\cite{liu2022pseudo,bao2022analytic,bao2022estimating,lu2022dpm,lu2023dpmsolver,zhao2023unipc} exist, we choose the popular DDIM in Eq.~(\ref{eq:sampling}) and tune its noise hyperparameter $\eta$.
The optimal value of $\eta$ depends on the base model and we search for the best one in $\{0, 0.5, 1\}$ as detailed in Tab.~\ref{tab:hyperparameters}. 

The whole process of SIR is presented in Algorithm~\ref{alg:sir}. Compared with existing methods~\cite{poole2022dreamfusion,wang2024prolificdreamer,liang2023luciddreamer,wu2023reconfusion} dedicated to enhancing 3D generation quality, SIR benefits from reconstruction and lowers the NFEs in total, consequently improving generation efficiency. (see Fig.~\ref{fig:compare-sds-nerf} for the comparison to SDS). 

\begin{figure}[t]
  \centering
  \includegraphics[width=1.0\linewidth]{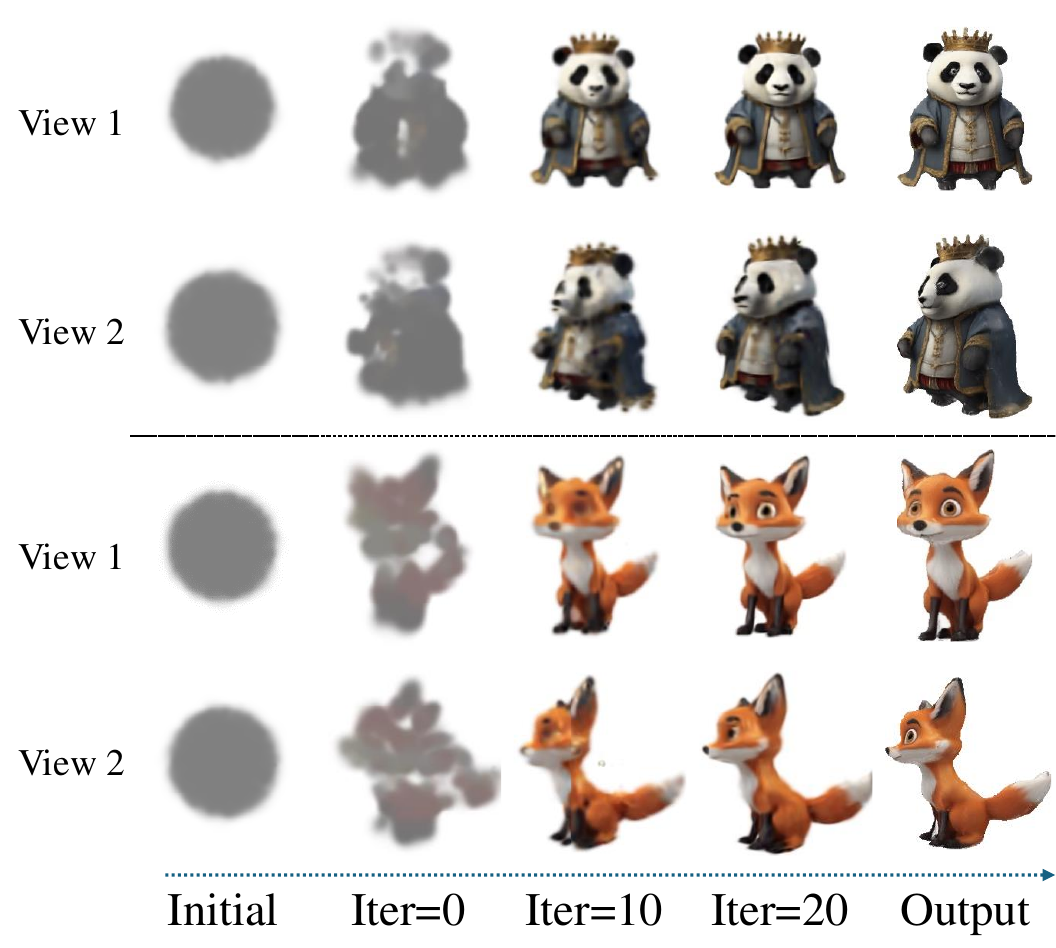}
  \caption{\textbf{Visualization of the optimization process in SIR}. The visual quality of the 3D samples increases along with the iterations.}
  \label{fig:optim-process}
  % \vspace{-.5cm}
\end{figure}

We visualize the optimization process of SIR in Fig.~\ref{fig:optim-process}. With the increase in the number of iterations, the visual quality of the 3D content improves, validating the effectiveness of our iterative reconstruction.

% After initialization, we employed the SIR algorithm to update the 3D parameters for multiple iterations. Moreover, an additional refinement step is performed to further improve the exported mesh texture's quality.

\subsection{Enabling optimization in pixel space}
\label{sec:sir-pixel-space}
Within the widely adopted framework of LDM~\cite{rombach2022high}, the rendered images are mapped through an encoder $\mathcal{E}$~\cite{kingma2014auto,van2017neural} to a latent space, where the loss function is calculated. Consequently, the gradients must be propagated back through the encoder, further reducing the efficiency of 3D object generation. In such a case, the SIR loss in latent space is given by:
\begin{align}
     \mathcal{J}_{\textrm{SIR-latent}}(\theta;c^{(k)}, \hat{x}^{(k)}) &= \| \mathcal{E}(g(\theta,c^{(k)}))-\hat{x}^{(k)}\|. \label{eq:sir-latent}
\end{align}

An alternative approach maps the diffusion output back to pixel space via the corresponding decoder $\mathcal{D}$, allowing for a similar loss function to be defined directly in pixel space. This method enables direct updates to 3D parameters without passing gradients through the encoder, thus enhancing efficiency. 

\begin{figure}[t]
    \centering
    \centering
    \begin{subfigure}{0.49\linewidth}
    \includegraphics[width=1.0\textwidth]{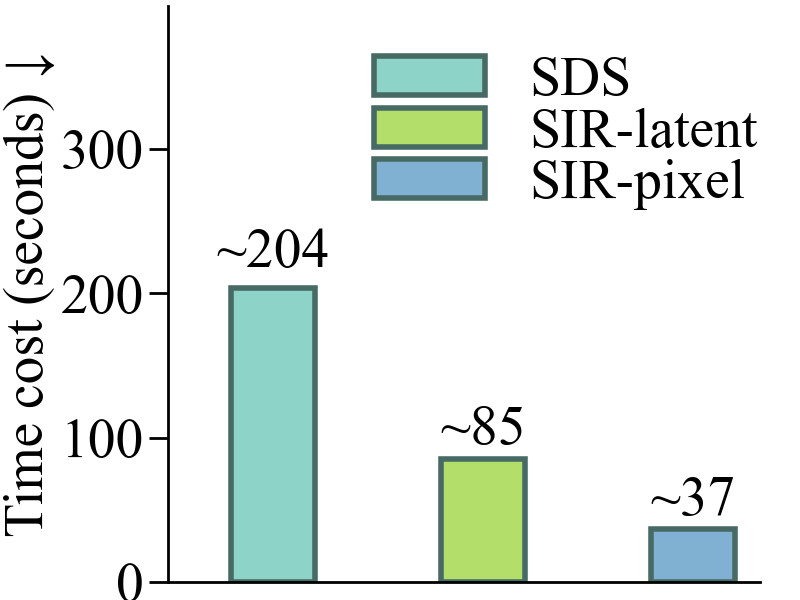}
    \caption{On Stable Zero123\cite{stable123}.}
    \label{fig:abla-1}
  \end{subfigure}
  % \hfill
  \begin{subfigure}{0.49\linewidth}
    \includegraphics[width=1.0\textwidth]{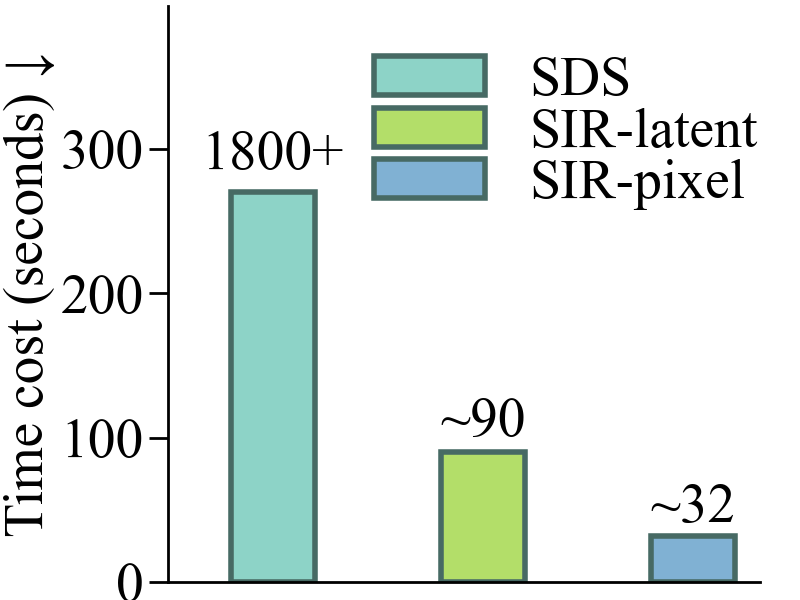}
    \caption{On ImageDream~\cite{wang2023imagedream}.}
    \label{fig:abla-2}
  \end{subfigure}
    \caption{\textbf{Optimization in pixel space accelerates generation}.}
    \label{fig:abla-optim-space}
\end{figure}

As discussed in Sec.~\ref{sec:sds-bottleneck}, previous optimization-based methods like SDS fail to be effective when applied in pixel space. In contrast, SIR achieves higher quality generation results through a carefully designed refinement process, thereby enabling optimization in pixel space. The SIR loss in pixel space is formalized as:
\begin{align}
     \mathcal{J}_{\textrm{SIR-pixel}}(\theta;c^{(k)}, \hat{x}^{(k)}) &= \| g(\theta,c^{(k)})-\mathcal{D}(\hat{x}^{(k)})\|, \label{eq:sir-pixel}
\end{align}
which is about 2-3 times faster than Eq.~(\ref{eq:sir-latent}) for 3D parameter optimization, as the analysis results shown in Fig.~\ref{fig:abla-optim-space}. It is set as the default loss throughout the paper. 

% From the ablation results in Fig.~\ref{fig:abla-1}, we can see that SIR-pixel achieves 2-3 times more acceleration than SIR-latent, verifying the benefits of SIR on enabling optimization in the pixel space. 

% However, we argue that this is not feasible for existing optimization methods~\cite{poole2022dreamfusion,wang2023score,wang2024prolificdreamer,liang2023luciddreamer,yu2023text,katzir2023noise} such as SDS. Even reparameterized as the data predicting form in Eq.~(\ref{eq:SDS-x0}), the pixel-level quality of the images obtained by SDS is poor for 3D optimization ({see Sec.~\ref{sec:abla-study}).

\section{MicroDreamer}
\label{sec:microdreamer}

We provide a general configuration in the 3D training and diffusion for the SIR algorithm. The comprehensive system is called \textbf{MicroDreamer} to highlight its efficiency. 

\emph{3D initialization and camera views.} We utilize a direct reconstruction approach to initialize the 3D content. Specifically, we optimize the loss function in Eq.~(\ref{eq:sir}) by several steps (see Tab.~\ref{tab:hyperparameters} for detailed values) to update the 3D parameters, where $\hat{x}$ represents the images sampled from random noise via the pre-trained multi-view diffusion models. The camera views are uniformly sampled following the corresponding baselines~\cite{tang2023dreamgaussian,threestudio2023} except the azimuth angles of different views in the same iteration are evenly distributed. In this setting, we are unable to directly utilize the diffusion model in~\cite{liu2023syncdreamer,long2023wonder3d,lu2023direct2,li2024era3d} due to their fixed camera views conditioned generation.
 
\emph{Annealed time schedule.} 
We utilize an annealed time schedule $\{t_2^{(k)}\}_{k=0}^{K-1}$ for the end of the forward process.  Intuitively, as the quality of the 3D assets improves, the input $x$ in Eq.~(\ref{eq:rendering}) becomes more proximate to the model distribution, thus requiring fewer steps to refine. This differs from SDS, which samples uniformly random $t_2$. Our preliminary experiments in Fig.~\ref{fig:abla2} suggest that a linearly annealed schedule is sufficient. The endpoints of the schedule depend on the diffusion model, detailed in Tab.~\ref{tab:hyperparameters}.

\begin{table}[t]
	\centering
 	% \caption{Design space of text-to-3D via 2D diffusion. We highlight the contributions of this paper that improve the fidelity, diversity and ability to generate complex scenes by $^{*}$, $^{\dagger}$ and $^{\ddagger}$ respectively.\label{table:framework}}
  \caption{\textbf{Key hyperparameters of MicroDreamer on three base diffusion models}. All models are employed to generate NeRF and the last two are employed to generate 3DGS and mesh. By default, the hyperparameters are shared across the two 3D representations. Otherwise, those for 3DGS are shown in brackets. $T$ is the end time of diffusion.}
  \label{tab:hyperparameters}
  \vspace{.15cm}
         \resizebox{\linewidth}{!}
         {% <------ Don't forget this %
	\begin{tabular}{lccc}
		\toprule
            Model Select & \textbf{MVDream}~\cite{shi2023mvdream}  & \textbf{Stable Zero123}~\cite{stable123} & \textbf{ImageDream}~\cite{wang2023imagedream}\\
        \midrule
        \textbf{Diffusion}&&&\\ 
        % Base model & Stable123 & Stable123 & Stable123 & Stable123\\
        {CFG}~\cite{ho2022classifier} & 7.5 & 3.0& 3.0 (2.0) \\
        Forward process & hybrid &hybrid & \makecell{hybrid}\\
        Time schedule of $t_2$ &$0.8T\rightarrow0.5 T$& \makecell{$0.8T\rightarrow 0.2T$ \\($0.9T\rightarrow 0.2T$)}& 
  \makecell{$0.8T \rightarrow0.6T$ \\ ($0.8T\rightarrow0.4T$) }  \\        
        Time schedule of $t_1$ & $t_1 = t^2_2/T$ & $t_1 = 0.6t_2$&  \makecell{$t_1 = t^2_2/T$ \\($t_1 = 0.6t_2$)} \\
        Sampling process & DDIM, $\eta=0.0$ & DDIM, $\eta=0.5$ & DDIM, $\eta=1.0$\\
        {Discretization steps} & 50 & 20 &10\\
        \midrule
        \textbf{3D training}&&&\\
        Resolution &$64\rightarrow128$ & \makecell{$64\rightarrow 128\rightarrow196$ \\ (256)} & \makecell{$64\rightarrow 128$ \\(256)} \\
        Background & learned by NN &always white&always white\\
        \# camera views &4&4 (6)&4\\
        {\# initialized steps} & 50 & 15 & 50\\
        {\# iterations $K$} & 50 & 30 (20 or 30) & 30 \\
        {\# reconstruction steps $I$} & 15 & 15 & 15\\
        Loss type & $\ell_1$&$\ell_1$&$\ell_1$\\
        Ref. color loss &- &0.1 (0.3) &0\\
        Ref. opacity loss &-&0.001 (0.01)&0\\
        \bottomrule
	\end{tabular}% <------ Don't forget this %
    }
    % \vspace{-.1in}
\end{table} 

\emph{Reference image loss for image-to-3D.} 
A reference image is available in image-to-3D, which is regarded as the ground truth front view of the 3D object on Stable Zero123~\cite{stable123} following DreamGaussian~\cite{tang2023dreamgaussian}. In this way, we add the reference loss in the same form as the reconstruction loss with Eq.~(\ref{eq:sir}) in each training iteration. The weight set for the reference loss can be seen in Tab.~\ref{tab:hyperparameters}.  

\emph{3DGS settings and mesh refinement.}
For simplicity, our approach on 3DGS largely adheres to the settings in DreamGaussian~\cite{tang2023dreamgaussian} unless specified. We incorporate a densify and prune procedure at every 100 updates during the initial 300 updates. At the end of optimization, we do a last prune and remove potential white Gaussians with scales larger than 0.01. We follow the mesh extraction method in LGM~\cite{tang2024lgm} and employ a threshold value of $5$ in the marching cube algorithm~\cite{lorensen1998marching}. We utilize SIR to optimize the exported mesh texture for one iteration with 30 steps, and we use a noise-adding process and DDIM with $\eta=0$ for simplicity.

\section{Experiments}
\label{sec:experiment} 

We present the experimental details and results on NeRF, 3DGS, and ablation sequentially.

\subsection{Experimental details}

We present key hyperparameters of MicroDreamer in Tab.~\ref{tab:hyperparameters}. For implementing SIR and SDS on NeRF, we choose the popularly used framework threestudio~\cite{threestudio2023}. For the implementation of SIR on 3DGS, we follow the framework from DreamGaussian~\cite{tang2023dreamgaussian}. Most hyperparameters are not sensitive to the results, and we follow the previous framework~\cite{threestudio2023,tang2023dreamgaussian} for these settings. For important hyperparameters, including the number of iterations, camera views, and the time schedule, see Sec.~\ref{sec:abla-study} for ablation. See Tab.~\ref{tab:codebase} for details about the URL of the codebases and checkpoints we used in this paper. 

% insensitive to most of the hyperparameters. 
% For important cfg. grid search xxx, l see xxx for ablation.

For a fair comparison, we utilize the widely used CLIP similarity~\cite{radford2021learning} for quantitative comparison following DreamGaussian~\cite{tang2023dreamgaussian}. We consider 8 views at 0 elevation and evenly distributed in azimuth angles starting from 0. For all methods, the corresponding images are rendered from NeRF in Sec.~\ref{sec:nerf-res} and mesh in Sec.~\ref{sec:3dgs-res}. We report the average CLIP similarities between these images and the reference image or text.

\begin{table}[tb]
	\centering
  \caption{\textbf{Codebases and checkpoints.} We provide URLs for the open-source assets we used in this paper.}
  \label{tab:codebase}
  \vspace{.15cm}
         \resizebox{1.0\linewidth}{!}
         {% <------ Don't forget this %
	\begin{tabular}{lc}
		\toprule
            Model &URL  \\
        \midrule
        \multicolumn{2}{l}{\textbf{Codebases}} \\
        \cite{threestudio2023} &\url{https://github.com/threestudio-project/threestudio} \\
        \cite{tang2023dreamgaussian} &\url{https://github.com/dreamgaussian/dreamgaussian} \\
        \midrule
        \multicolumn{2}{l}{\textbf{Checkpoints}} \\
        \cite{cherti2022reproducible}  & \url{https://huggingface.co/laion/CLIP-ViT-bigG-14-laion2B-39B-b160k}\\
          \cite{shi2023mvdream} &\url{https://github.com/bytedance/MVDream-threestudio}\\
          \cite{stable123} &\url{https://huggingface.co/stabilityai/stable-zero123}\\
          \cite{wang2023imagedream} &\url{https://github.com/bytedance/ImageDream}\\
          \midrule
          \multicolumn{2}{l}{\textbf{Baselines}} \\
          \cite{nichol2022point} &{\url{https://github.com/openai/point-e}}\\
          \cite{jun2023shap}& {\url{https://github.com/openai/shap-e}}\\
          \cite{liu2023one} &{\url{https://github.com/One-2-3-45/One-2-3-45}} \\
          \cite{zou2023triplane}& {\url{https://github.com/VAST-AI-Research/TriplaneGaussian}}\\
          \cite{long2023wonder3d} &{\url{https://github.com/xxlong0/Wonder3D}}\\
          \cite{tang2024lgm}& {\url{https://github.com/3DTopia/LGM}}\\
          \cite{openlrm}& {\url{https://github.com/3DTopia/OpenLRM}}\\
        \bottomrule
	\end{tabular}% <------ Don't forget this %
    }
    \vspace{-.1in}
\end{table}

\begin{figure*}[tb]
    \centering
  \begin{subfigure}{0.3\linewidth}
  \centering
    \includegraphics[width=1.0\textwidth]{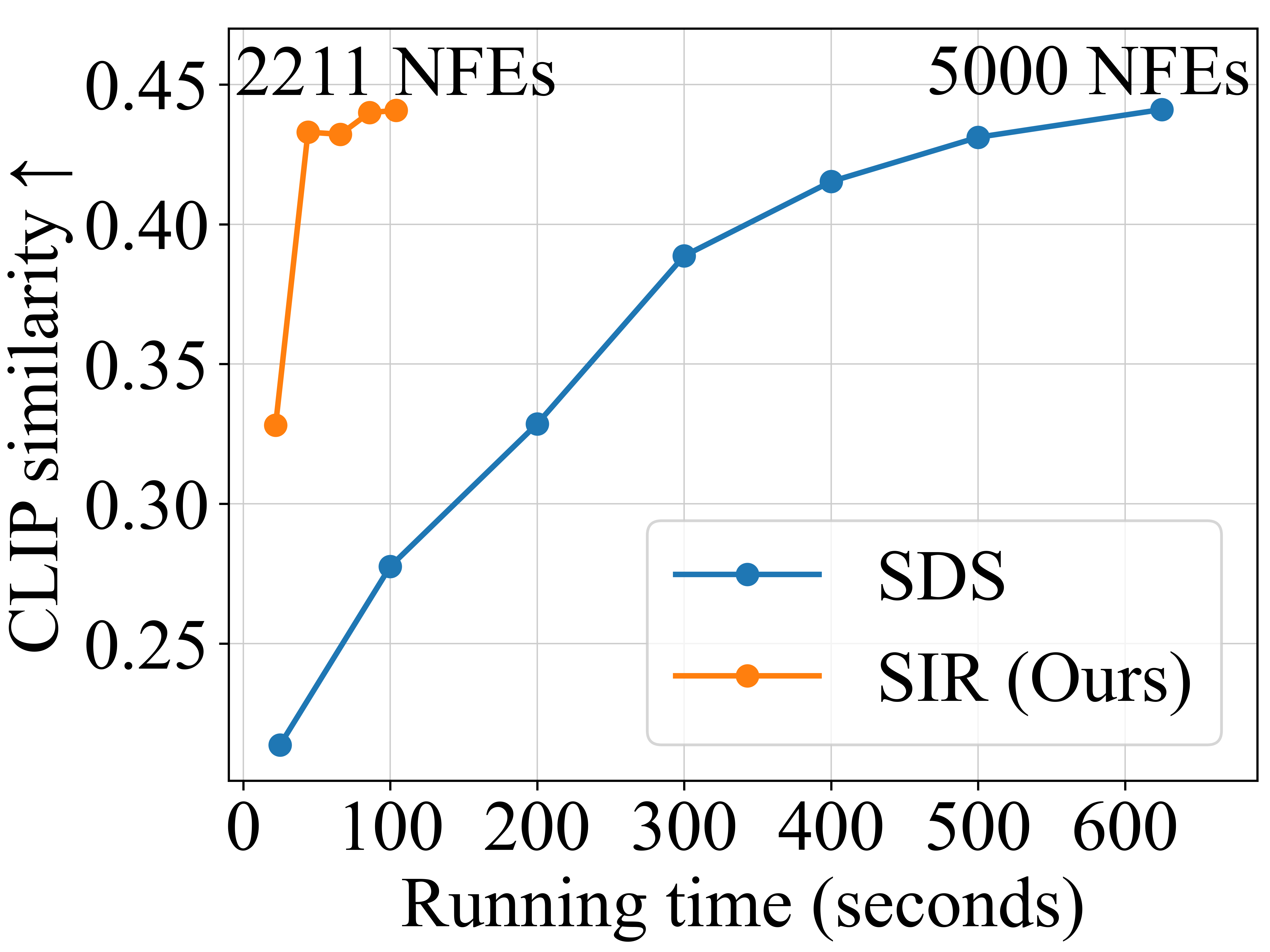}
    \caption{Comparion on MVDream~\cite{shi2023mvdream}}
    \label{fig:comp-sds-mvdream}
  \end{subfigure}
  % \hfill
  \begin{subfigure}{0.3\linewidth}
  \centering
    \includegraphics[width=1.0\textwidth]{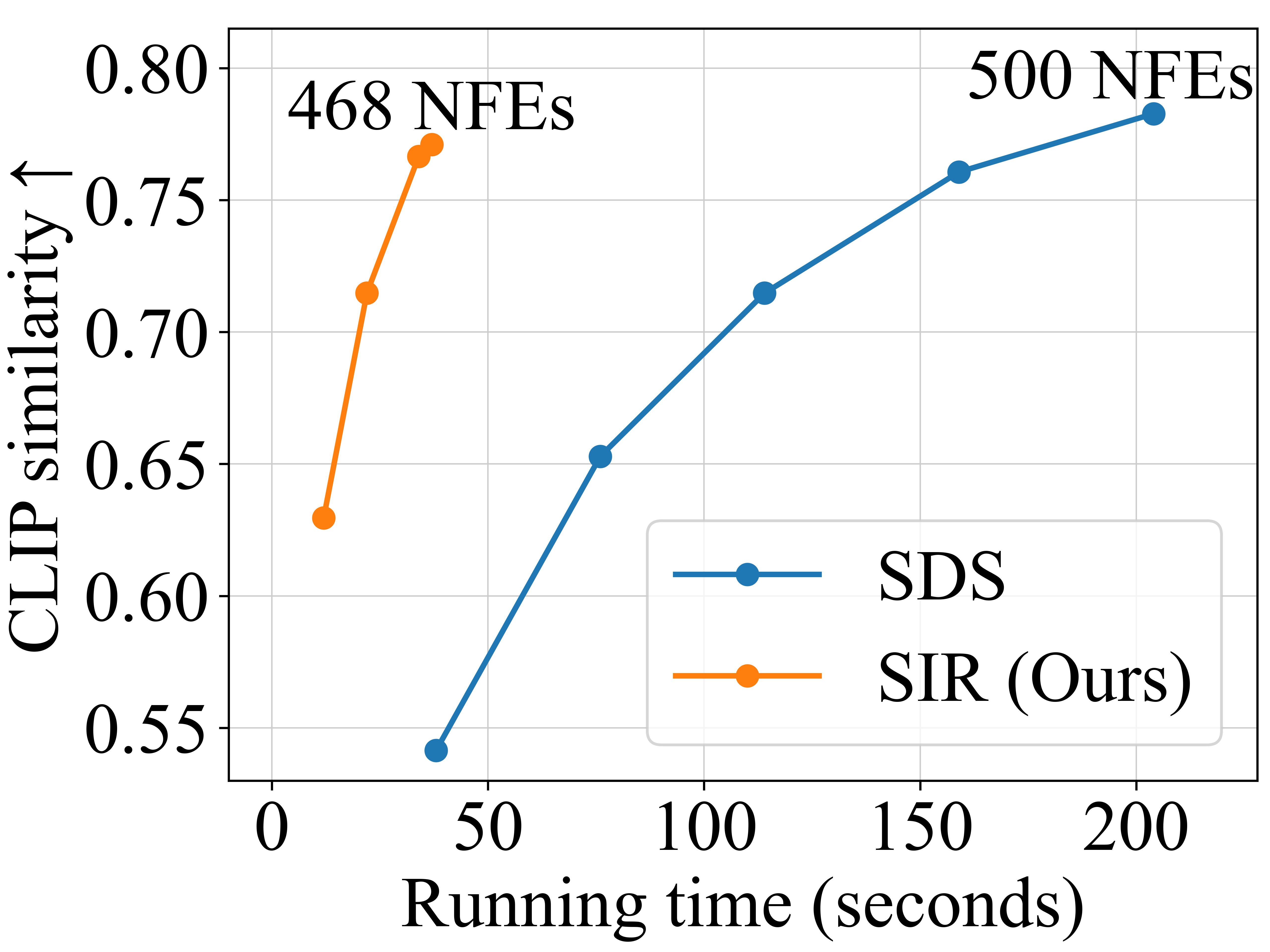}
    \caption{Comparion on Stable Zero123~\cite{stable123}}
    \label{fig:comp-sds-stable123}
  \end{subfigure}
  % \hfill
  \begin{subfigure}{0.3\linewidth}
  \centering
    \includegraphics[width=1.0\textwidth]{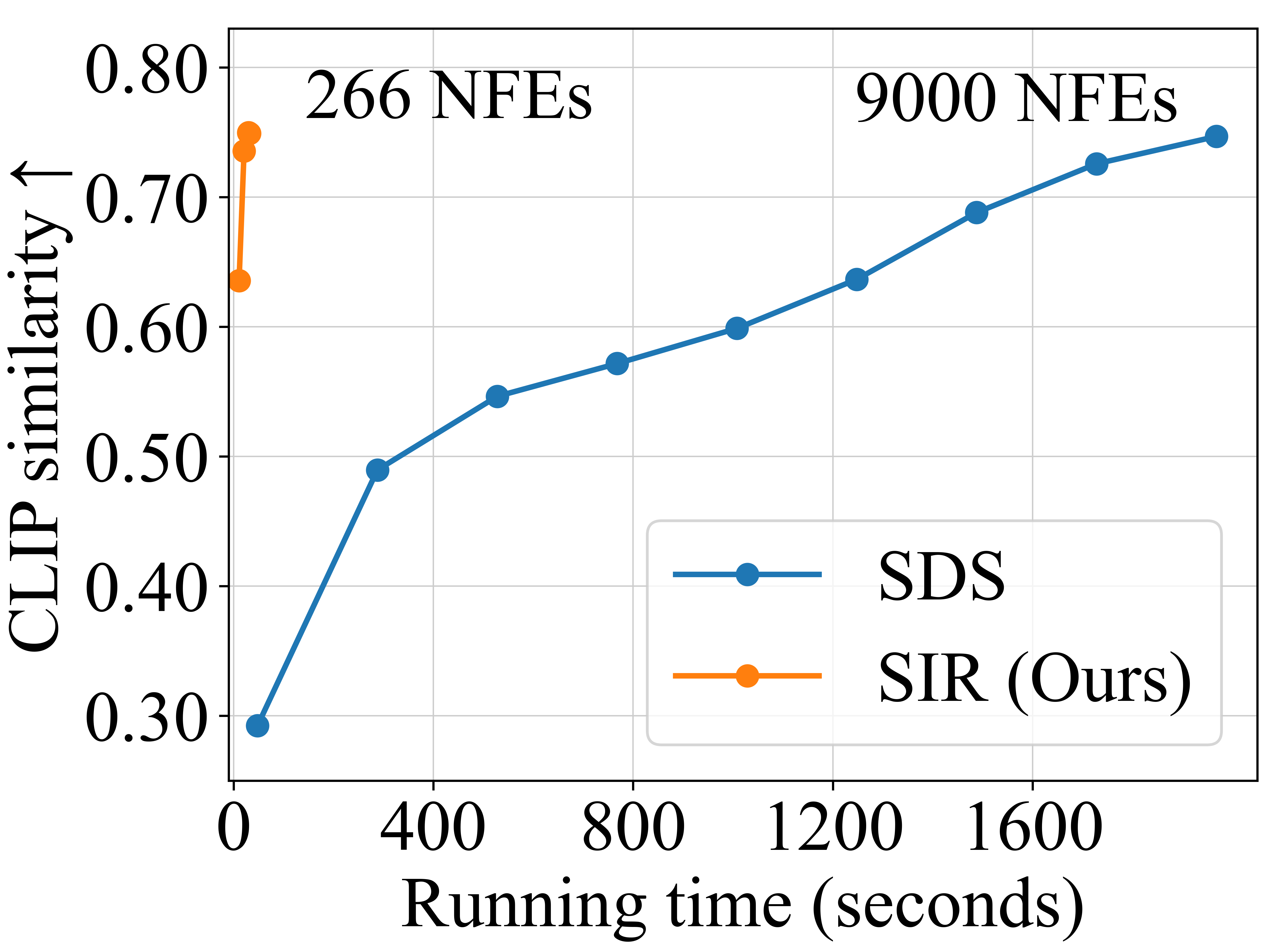}
    \caption{Comparion on ImageDream~\cite{wang2023imagedream}}
    \label{fig:comp-sds-imagedream}
  \end{subfigure}
  \caption{\textbf{Comparison of SIR and SDS on NeRF.} We plot the curve of the CLIP similarity in the generation process and final NFEs on different models. SIR lowers the NFEs and achieves a 5-20 times acceleration to achieve a competitive quality.}
  \label{fig:compare-sds-nerf}
\end{figure*}

\begin{figure}[t]
    \centering
    \begin{subfigure}{1.0\linewidth}
      \includegraphics[width=1.0\textwidth]{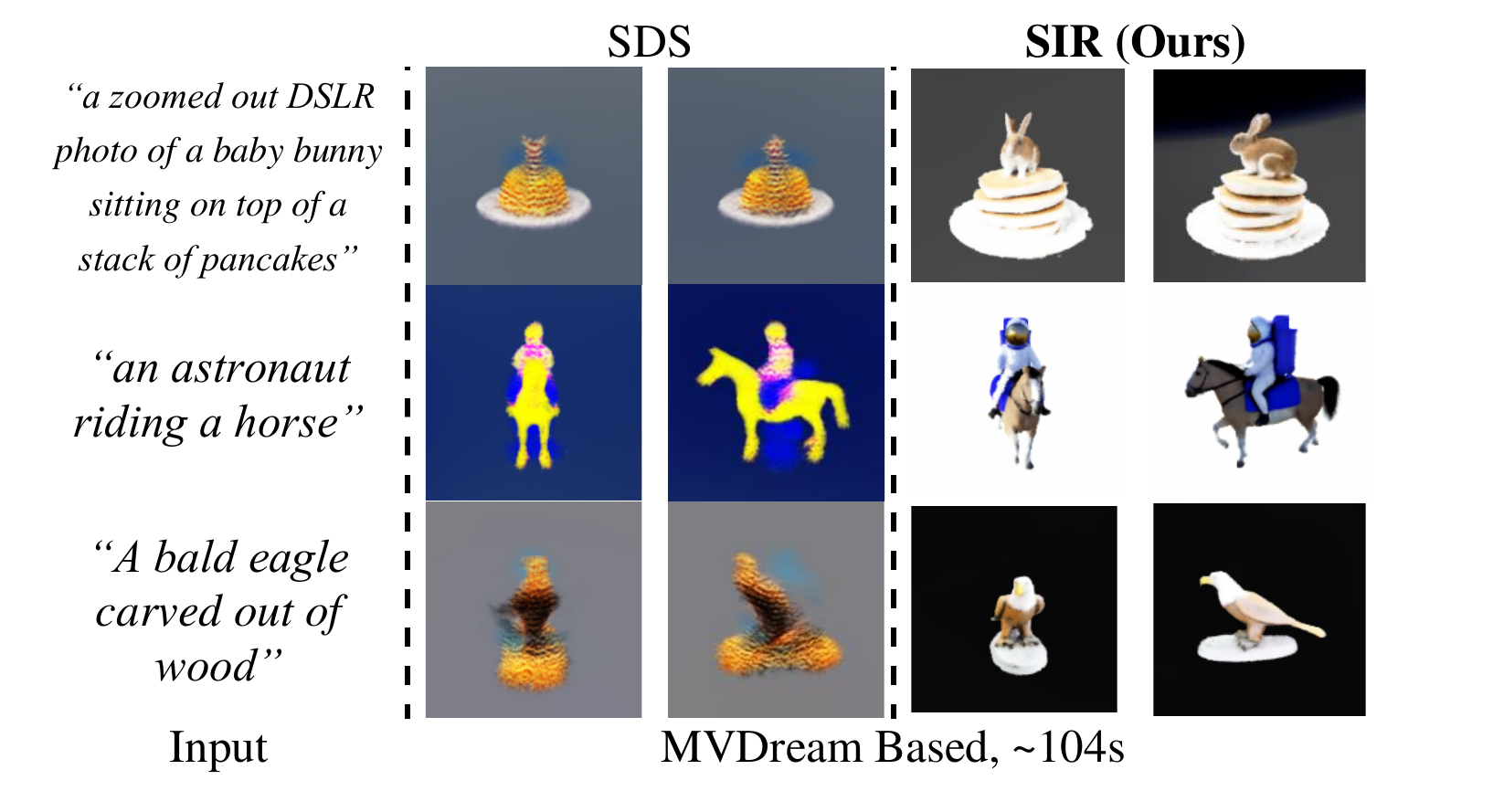}
      % \caption{{Results for MVDream~\cite{shi2023mvdream} on NeRF.}}
      \label{fig:nerf-mv}
  \vspace{-.5cm}
    \end{subfigure}
    \begin{subfigure}{0.9\linewidth}
      \centering
      \includegraphics[width=1.0\textwidth]{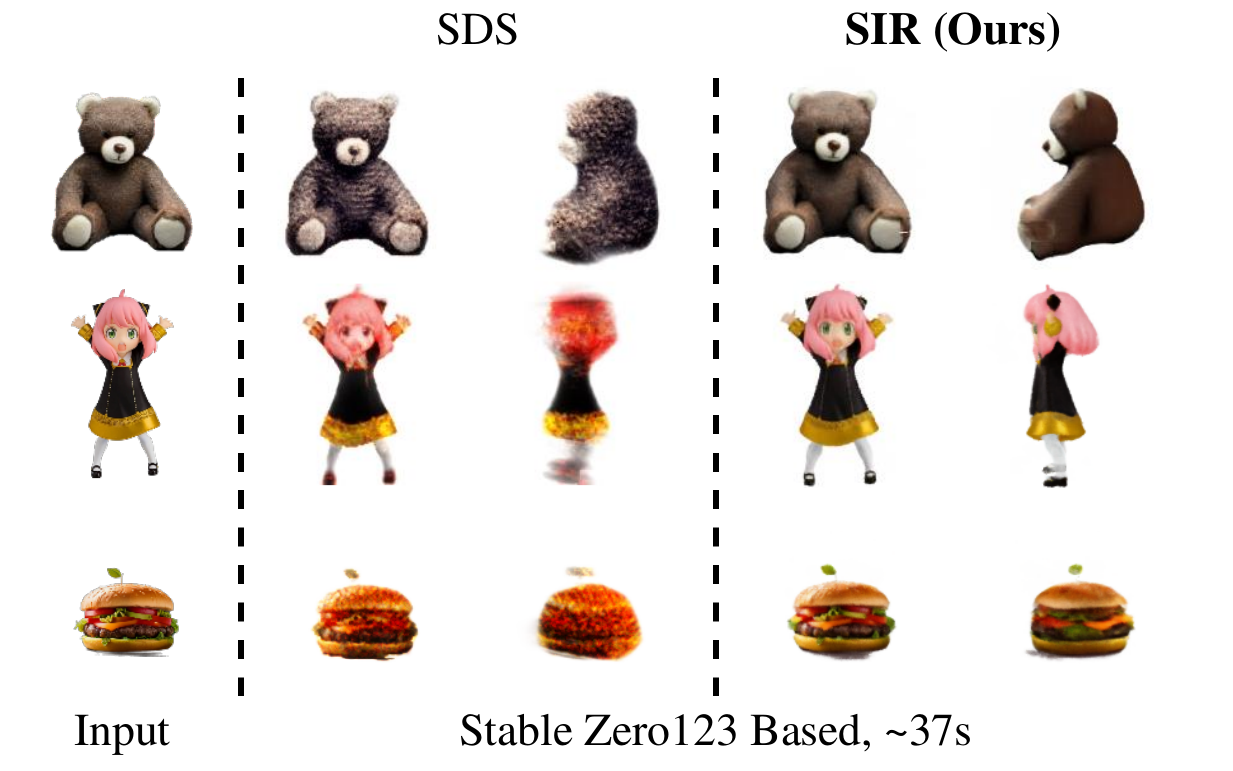}
      % \caption{{Results for Stable Zero123~\cite{stable123,liu2023zero1to3} on NeRF.}}
      \label{fig:nerf-sai}
      \vspace{-.5cm}
    \end{subfigure}
    
    \begin{subfigure}{0.9\linewidth}
      \centering
      \includegraphics[width=1.0\textwidth]{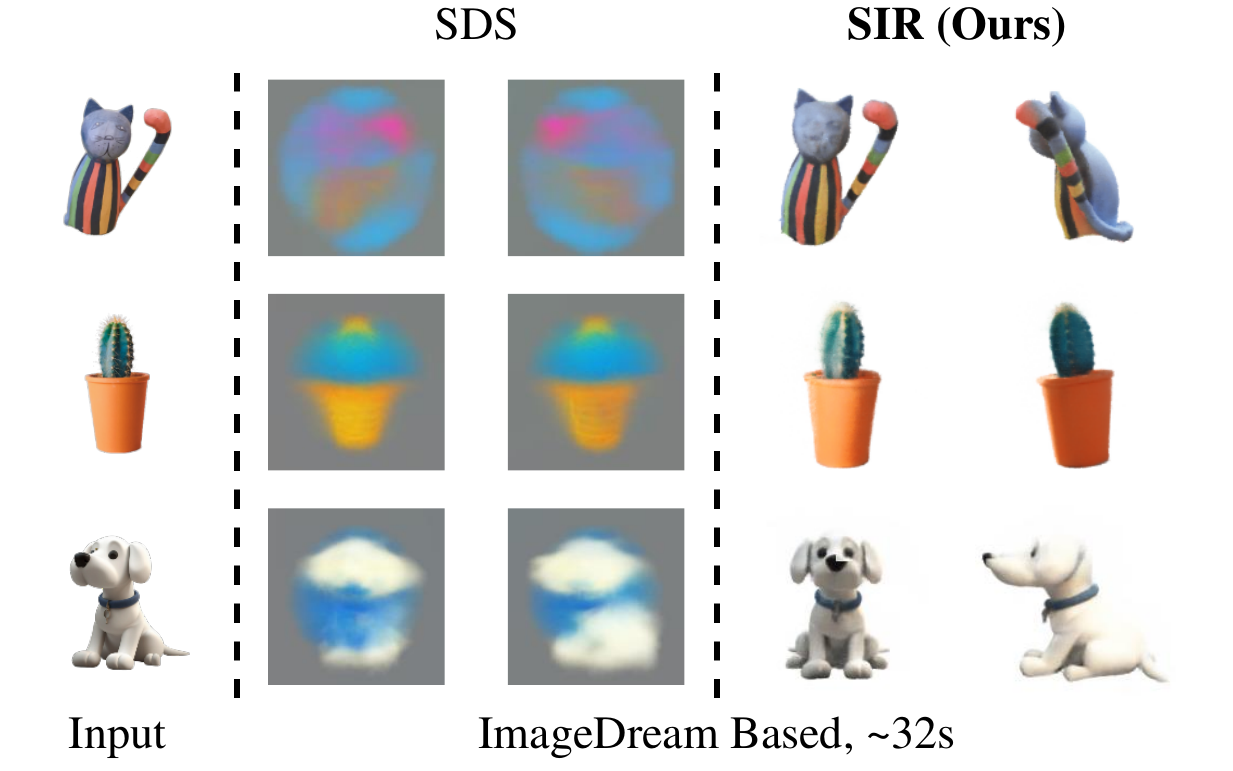}
      % \caption{{Results for ImageDream~\cite{wang2023imagedream} on NeRF.}}
      \label{fig:nerf-ip}
      \vspace{-.5cm}
    \end{subfigure}
    \caption{\textbf{Qualitative comparison on NeRF.} SIR can generate NeRF of higher visual quality than SDS in a short time.}
    \label{fig:nerf-qual}
    \vspace{-.5cm}
\end{figure}

\subsection{Results on NeRF}
\label{sec:nerf-res} 

We apply SIR algorithm on NeRF~\cite{mildenhall2021nerf,muller2022instant,threestudio2023} leveraging three multi-view diffusion models from MVDream~\cite{shi2023mvdream}, Stable Zero123~\cite{stable123}, and ImageDream~\cite{wang2023imagedream}. For each model, we selected 6 input prompts from a widely used codebase~\cite{threestudio2023} for testing, calculated NFEs, and recorded the average CLIP similarity during the generation process. Compared with SDS, SIR lowers the total NFEs and accelerates the generation process 5-20 times while holding a competitive CLIP similarity, as shown in Fig.~\ref{fig:compare-sds-nerf}. 

To provide a comprehensive comparison, we qualitatively analyze the generation results of SDS and SIR under identical time conditions, as illustrated in Fig.~\ref{fig:nerf-qual}. Notably, at the moment of convergence in SIR, we find that SDS has not achieved satisfactory results, such as producing ambiguous outcomes. These observations support the quantitative results, demonstrating that SIR is general and more efficient.

% The results show that SIR can generate NeRFs of higher visual quality than SDS. ??

% 保持相同CLIP，xxx across all base models. effective and general.

% For a complete comparison, we xxx qualitatively. 直观地比较，固定了一个相同的时间，比较视觉效果。In particular, 我们考虑SIR 收敛的时间，发现SDS 仍未取得令人满意的结果，比如非常模糊xx。综上所述， efficient 方法，agree with the quantitative results.

\subsection{Results on 3D Gaussian splatting}
\label{sec:3dgs-res}

\begin{figure}[t]
  \centering
  \includegraphics[width=1.0\linewidth]{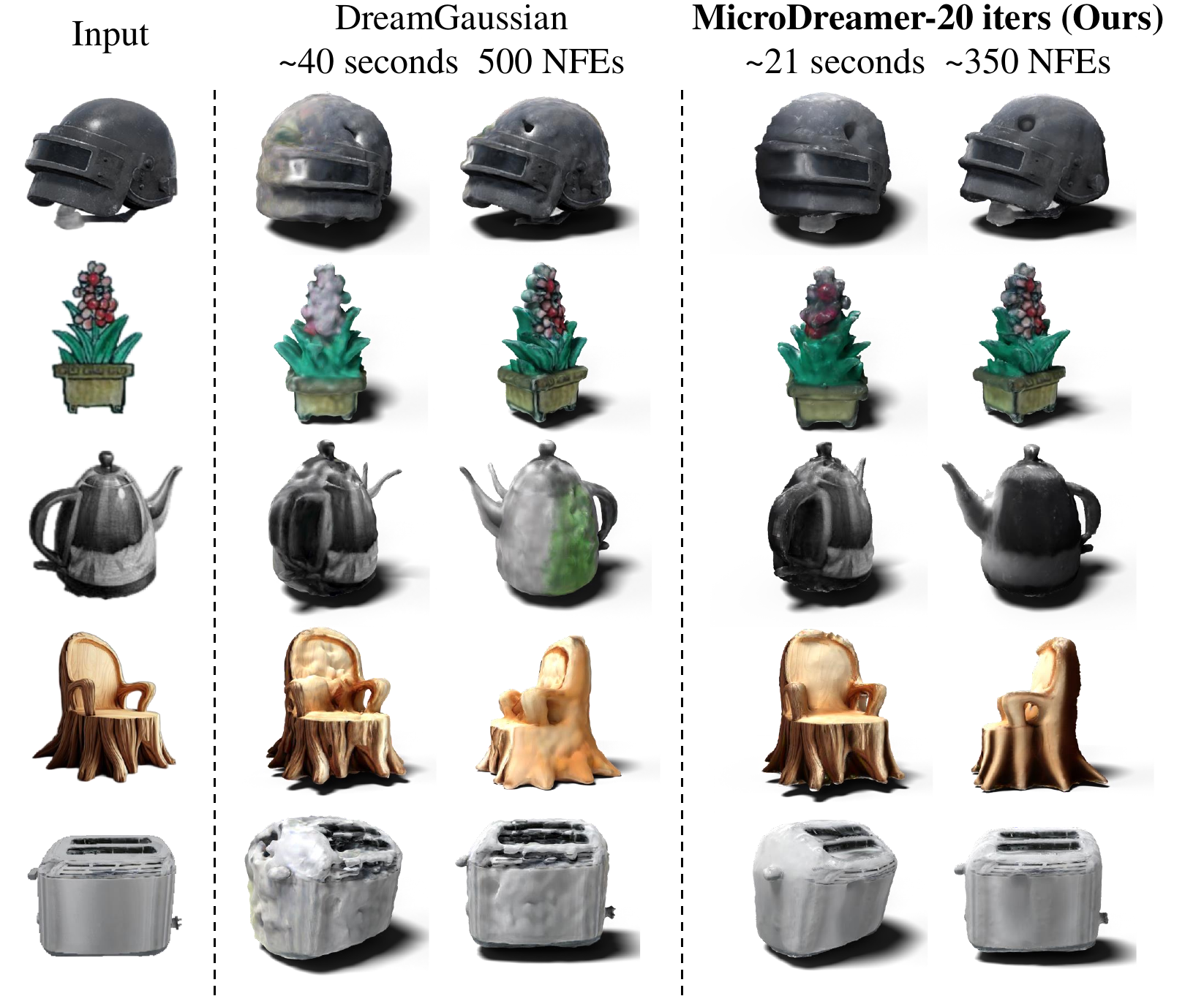}
  \caption{\textbf{Qualitative comparisons on mesh from 3DGS.} MicroDreamer produces superior meshes, characterized by enhanced texture and reduced geometric artifacts, with greater efficiency than DreamGaussian.
  }
  \label{fig:qual-comp}
\end{figure}

\emph{Qualitative comparisons.}
In Fig.~\ref{fig:qual-comp}, we present the generated 3D meshes comparing the fastest optimization-based baseline, DreamGaussian~\cite{tang2023dreamgaussian}, with our method. MicroDreamer reduces DreamGaussian's generation time by half while exhibiting superior performance, including improved texture and enhanced geometric structure.

% 根据图描述的具体一些。图对应描述，突出一下不一样的地方。润色一下语言。

% MicroDreamer achieves better meshes (e.g., better texture and fewer artifacts on the geometry) with higher efficiency than DreamGaussian.

\emph{Quantitative comparisons.}
We compare MicroDreamer with eight competitive baselines including Point-E~\cite{nichol2022point}, Shap-E~\cite{jun2023shap}, One-2-3-45~\cite{liu2023one}, TriplaneGaussian\footnote{As TriplaneGaussian has no official mesh export code, we apply the mesh exported code from LGM~\cite{tang2024lgm} for it.}~\cite{zou2023triplane}, Wonder3D~\cite{long2023wonder3d}, LGM~\cite{tang2024lgm}, Open-LRM~\cite{openlrm} and DreamGaussian~\cite{tang2023dreamgaussian}. We record the generation time for each model on a single NVIDIA A100 (80GB) GPU and compute the average CLIP similarity for mesh on a test dataset consisting of 87 images collected from previous works~\cite{liu2023one,liu2023syncdreamer,tang2023dreamgaussian}, as shown in Tab.~\ref{tab:quan-comp}. MicroDreamer generates significantly better 3D content than DreamGaussian, as indicated by the standard deviation across multiple runs, and has higher efficiency.

In addition, MicroDreamer is on par with speed compared to feed-forward methods~\cite{openlrm} trained on a substantial amount of 3D data and has a very competitive CLIP similarity. In conclusion, all those results suggest that SIR is a promising approach for efficient 3D generation.
% Notably, the sample quality of MicroDreamer can be improved as the multi-view diffusion evolves. See detailed limitations and future work in Sec.~\ref{sec:conclusion}.

\begin{table}[tb]
	\centering
  \caption{\textbf{Quantitative comparisons.} MicroDreamer significantly outperforms the strong optimization-based baseline DreamGaussian in quality and efficiency and remains competitive with feed-forward methods. All results are averaged over three runs.}
  \label{tab:quan-comp}
  \vspace{.15cm}
         \resizebox{1.0\linewidth}{!}
         {% <------ Don't forget this %
	\begin{tabular}{lcc}
		\toprule
            Method & CLIP similarity $\uparrow$& Generation time $\downarrow$\\
        \midrule
        Point-E~\cite{nichol2022point} &  $0.566\pm 0.0011$&$\sim 24$s\\
        Shap-E~\cite{jun2023shap}& $0.626 \pm 0.0030$&$\sim 5$s\\ 
        One-2-3-45~\cite{liu2023one} & $0.617 \pm 0.0025$&$\sim 42$s\\ 
        Wonder3D~\cite{long2023wonder3d} & $0.696 \pm 0.0017$&$\sim 170$s\\ 
        TriplaneGaussian~\cite{zou2023triplane} &  $0.691\pm 0.0016$&$\sim 7$s\\  
        LGM~\cite{tang2024lgm} & $0.700 \pm 0.0017$&$\sim 3$s \\  
        Open-LRM-mix-base-1.1\cite{openlrm} & $0.704 \pm 0.0000$&$\sim 23$s \\  
        % TripoSR & 0.723&$< 1$s \\  
        \midrule
        DreamGaussian~\cite{tang2023dreamgaussian}&$0.692 \pm 0.0015$&$\sim 30+10$s\\ 
        DreamGaussian-300 iter~\cite{tang2023dreamgaussian} &$0.641 \pm 0.0032$&$\sim 18+10$s\\ 
        MicroDreamer-20 iter (\textbf{Ours})&$0.697\pm 0.0009$ &$\sim 18+3$s\\
        MicroDreamer-30 iter (\textbf{Ours})&$0.711\pm 0.0007$&$\sim 26+3$s\\
        \bottomrule
	\end{tabular}% <------ Don't forget this %
    }
    % \vspace{-.1in}
\end{table}

\subsection{Analysis and ablation study}
\label{sec:abla-study}
% \todo{analysis or ablation}

% \begin{figure}[t]
%   \centering
%   \begin{subfigure}{0.325\linewidth}
%     \includegraphics[width=1.0\textwidth]{appendix_fig/A-1.png}
%   \end{subfigure}
%   \hfill
%   \begin{subfigure}{0.325\linewidth}
%     \includegraphics[width=1.0\textwidth]{appendix_fig/A-2.png}
%   \end{subfigure}
%   \hfill
%   \begin{subfigure}{0.325\linewidth}
%     \includegraphics[width=1.0\textwidth]{appendix_fig/A-3.png}
%   \end{subfigure}
%   % \hfill
%   % \begin{subfigure}{0.24\linewidth}
%   %   \includegraphics[width=1.0\textwidth]{appendix_fig/A-4.png}
%   % \end{subfigure}
%   \caption{\textbf{Results of single-step prediction for SDS.} The left is the input and the right is the output of single-step prediction for each pair of images. The poor quality of the output makes it unsuitable for 3D optimization.}
%   \label{fig:single-pred}
% \end{figure}

% \begin{figure}[t]
%     \centering
%     \includegraphics[width=1.0\linewidth]{appendix_fig/A.SDS.pdf}
%     \caption{\textbf{Results of applying SDS in pixel space.} SDS optimizing in pixel space can not be effective.}
%     \label{fig:sds-pixel-res}
% \end{figure}

\begin{figure}[t]
  \centering
  \begin{subfigure}{1.0\linewidth}
    \includegraphics[width=1.0\textwidth]{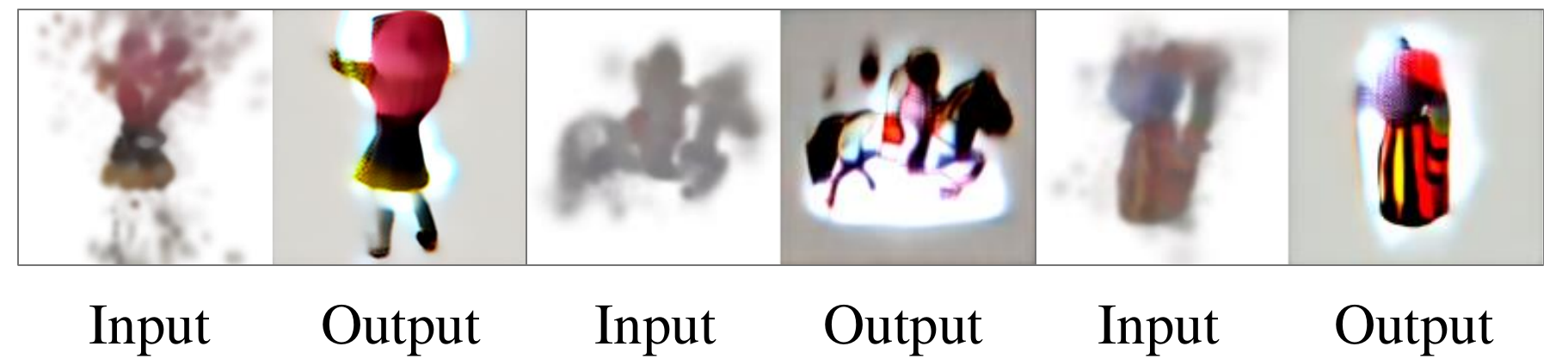}
    \caption{Results of single-step prediction using Eq.~(\ref{eq:x0t}) for SDS.}
  \end{subfigure}
  \hfill
  \begin{subfigure}{1.0\linewidth}
    \includegraphics[width=1.0\textwidth]{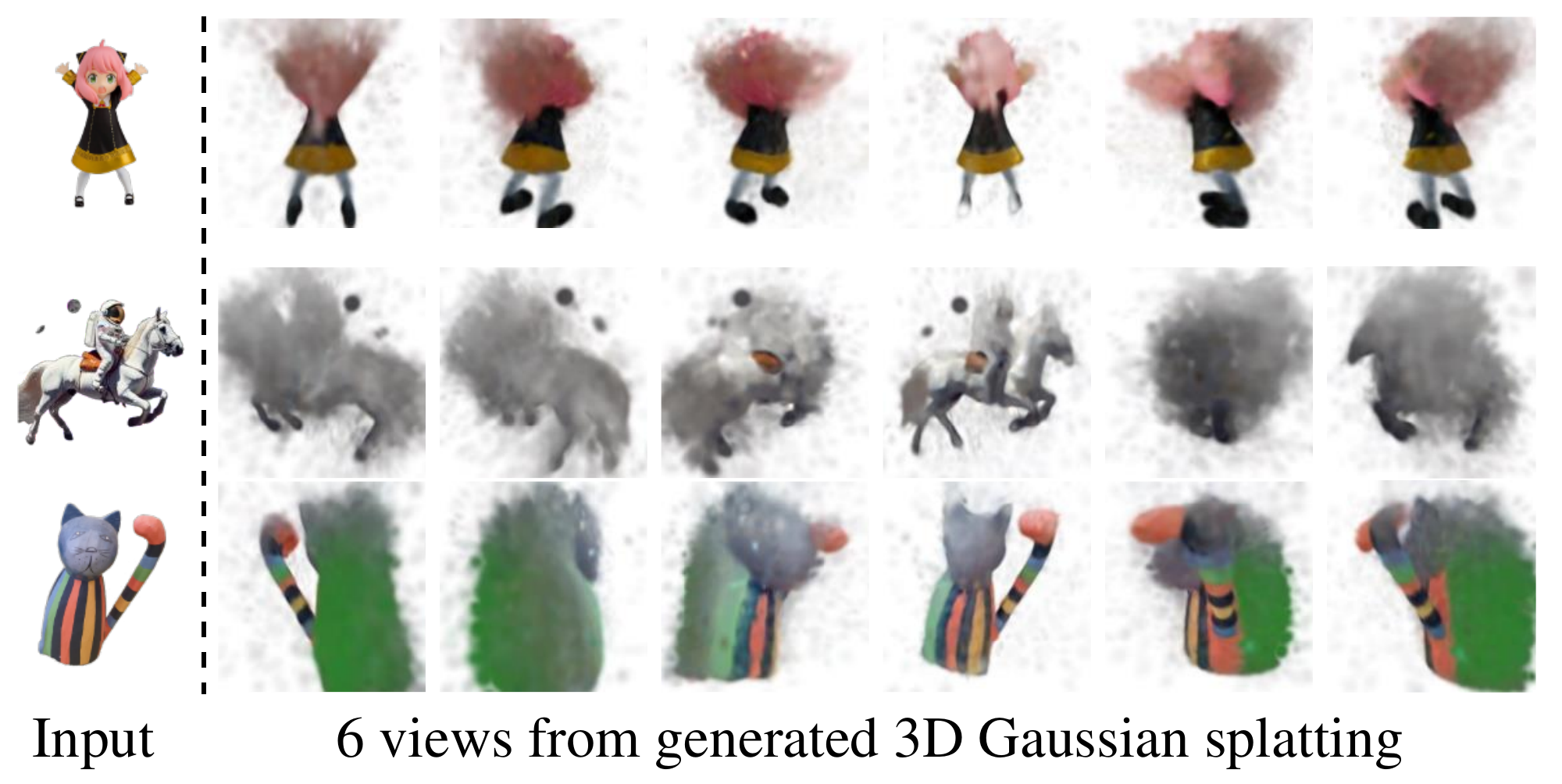}
    \caption{Results of applying SDS in pixel space.}
  \end{subfigure}
  \caption{\textbf{Failure of applying SDS in pixel space.} SDS optimizing in pixel space generates poor results.}
  \label{fig:sds-fail}
\end{figure}

\begin{figure}[t]
    
        \begin{subfigure}{0.49\linewidth}
            \includegraphics[width=1.0\textwidth]{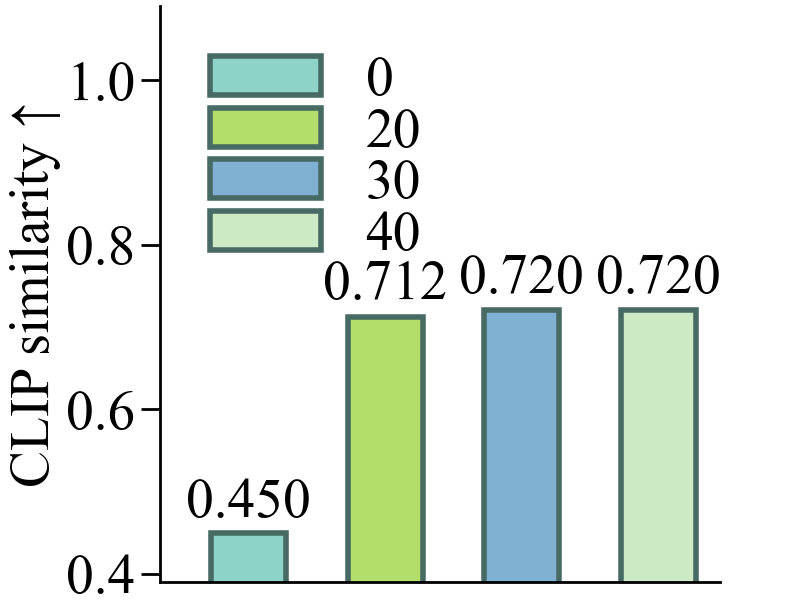}
            \caption{Iterative reconstruction.}
            \label{fig:abla0}
        \end{subfigure}
  \hfill
  \begin{subfigure}{0.49\linewidth}
    \includegraphics[width=1.0\textwidth]{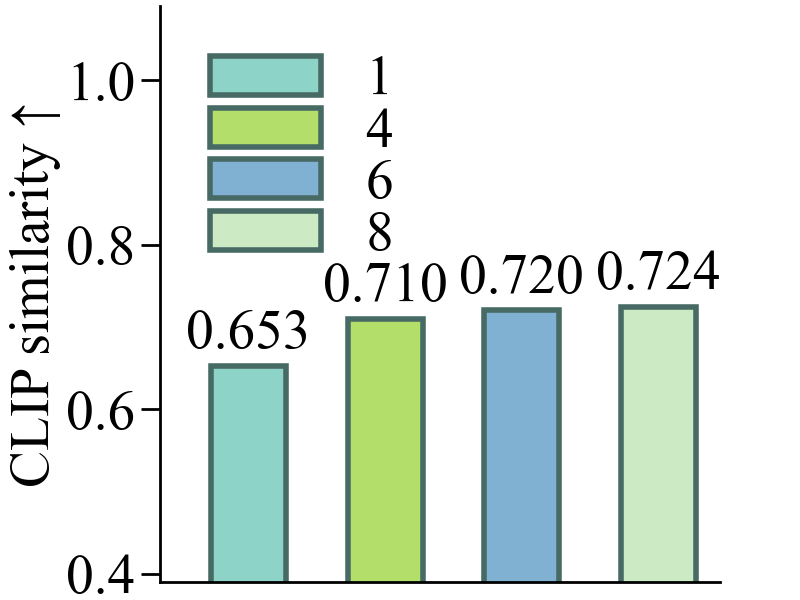}
          \caption{Camera views.}
    \label{fig:abla1}
  \end{subfigure}
  \hfill
  \begin{subfigure}{0.49\linewidth}
    \includegraphics[width=1.0\textwidth]{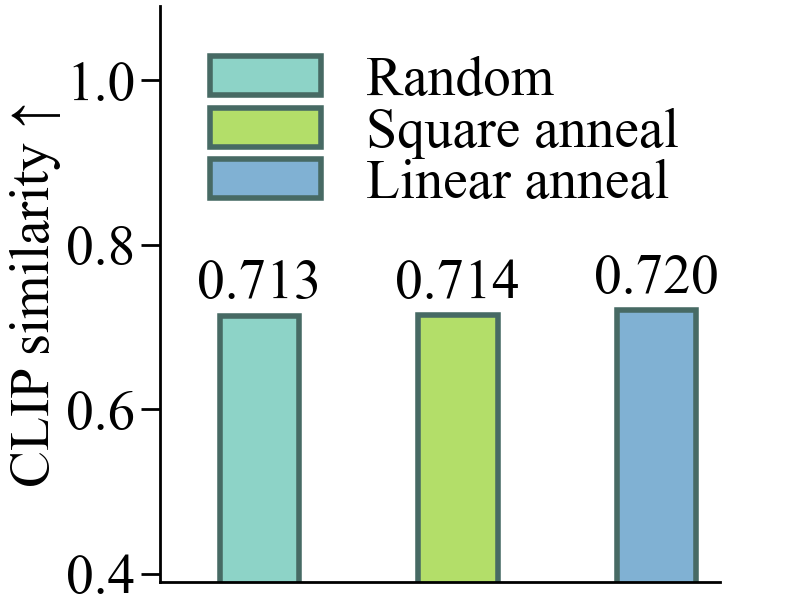}
        \caption{Time schedule.}
    \label{fig:abla2}
  \end{subfigure}
  \begin{subfigure}{0.49\linewidth}
    \includegraphics[width=1.0\textwidth]{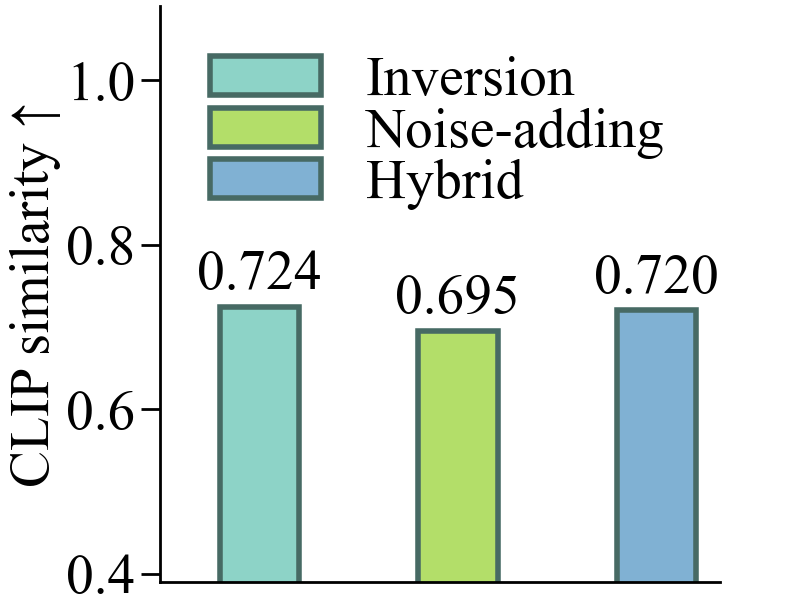}
    \caption{Forward process.}
    \label{fig:abla3}
  \end{subfigure}
    \caption{\textbf{Detailed analyses.} (a) Iterative reconstruction is necessary. (b) More camera views benefit. (c)  A linear schedule is sufficient. (d) A hybrid forward process can be effective.}
    \label{fig:abla-study}
\end{figure}

\emph{Analysis on optimization space.}
We implement SIR-latent on NeRF and compare its time consumption with SIR-pixel, as illustrated in Fig.~\ref{fig:abla-study}a-b. We can see that SIR-pixel achieves 2-3 times more acceleration than SIR-latent, verifying the benefits of SIR in enabling optimization in pixel space. 
Contrary to SIR, we argue that SDS optimizing in pixel space is not effective even if it has the data predicting form by reparameterization as presented in Eq.~({\ref{eq:SDS-x0}}). If we attempt to employ SDS for optimization in pixel space, i.e. optimizing 3D objects with 2D images like those generated by Eq.~({\ref{eq:x0t}}) in Fig.~\ref{fig:sds-fail}a, the resulting 3D content would be of subpar quality, as the results shown in Fig.~\ref{fig:sds-fail}b.

% Fig.~\ref{fig:sds-fail}(a) shows the poor quality for single-step prediction from the diffusion model using Eq.~({\ref{eq:x0t}}). Therefore, 

\emph{Ablation on iterative reconstruction.}
Fig.~{\ref{fig:abla0}} shows the necessity of iterative reconstruction. The case of $K=0$ outputs our 3D initialization, which is reconstructed by a single process from images generated by random noise. When $K>30$, the generation quality doesn't improve significantly. For efficiency, we set $K = 20$ or $30$. 

\emph{Ablation on number of camera poses.}
SIR necessitates more than 1 camera pose for reconstruction, as shown in Fig.~\ref{fig:abla1}, and the performance increases with more camera poses. We choose $6$ on 3DGS with Stable Zero123~\cite{stable123} model to balance efficiency and quality. See Tab.~\ref{tab:hyperparameters} for values in other settings.

\emph{Ablation on time schedule.}
We compare our linear schedule, the random schedule employed in SDS~\cite{poole2022dreamfusion}, and a square schedule with $t_2^{(k)}=(\frac{k}{K})^2\cdot(0.9T-0.2T)+0.2T$. As shown in Fig.~\ref{fig:abla2}, the schedules perform similarly. For simplicity and efficiency, we use the linear schedule by default.

\emph{Ablation on forward process.}
Fig.~\ref{fig:abla3} compares three different forward processes on 3DGS with the Stable Zero123~\cite{stable123} model. The hybrid strategy performs better than noise-adding. As discussed in Sec.~\ref{sec:refine}, we adopt the hybrid process rather than inversion for efficiency.

% \subsection{Failure of applying SDS in pixel space.}
% \label{apd:sds-pixel}

\subsection{Limitation}

\begin{figure}[t]
    \centering
    \includegraphics[width=1.0\linewidth]{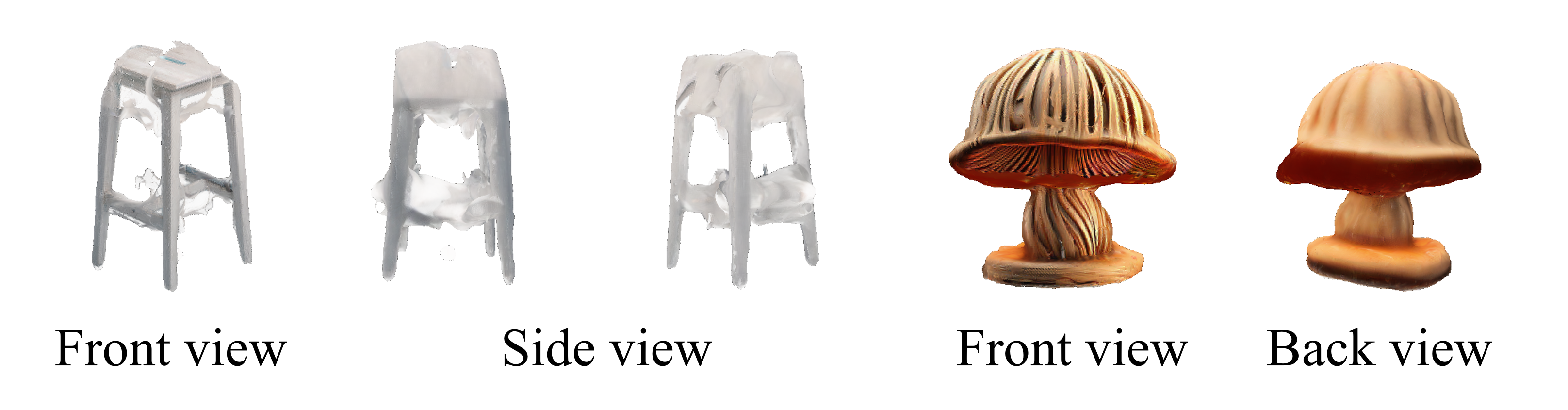}
    \caption{\textbf{Limitation.} Visualization of some less satisfactory cases generated by MicroDreamer.}
    \label{fig:fail}
\end{figure}

We show some less successful cases generated by our method in Fig.~\ref{fig:fail}. MicroDreamer faces challenges in generating complex geometric structures such as central hollows and may produce meshes with poor back surface textures. These problems may be mitigated as the quality of the generated multi-view images improves.

\section{Conclusion}
\label{sec:conclusion}

% Conclusions should not summarize but instead outline future goals or lessons learned.

We introduce SIR, an efficient and general algorithm combining the strengths of 3D reconstruction and iterative optimization to reduce total NFEs and enable optimization in pixel space in optimization-based 3D generation. SIR achieves a 5 to 20 times speed increase in NeRF generation compared to SDS. Remarkably, MicroDreamer generates high-quality meshes from 3DGS in about 20 seconds, outpacing the fastest optimization-based baseline DreamGaussian in quality and efficiency, and matching the speed of some feed-forward approaches with a competitive generation quality. 

There is potential for further improving MicroDreamer's efficiency via employing consistency models~\cite{song2023consistency,luo2023latent} or alternative sampling models that require fewer steps~\cite{yin2023one,sauer2023adversarial}. Additionally, the fidelity and 3D consistency of the objects produced by MicroDreamer are directly limited by the quality of the outputs from multi-view diffusion. Nevertheless, we believe SIR is promising and may inspire future work as the multi-view diffusion evolves. 

Furthermore, MicroDreamer can provide artists with the convenience of creating 3D assets. However, as a generative approach, our method may also be used for fabricating data and news. Moreover, the 3D content generated from input images may infringe on the privacy or copyright of others. While automatic detection might mitigate these issues, they warrant careful consideration.

% \todo{caption!!! message. move fig to intro. add std (if added. introduction, abs significantly compared to the std of the measurement) experiment method.}

\bibliographystyle{IEEEtran}
\bibliography{main}

\begin{IEEEbiography}[{\includegraphics[width=1in,height=1.25in,clip,keepaspectratio]{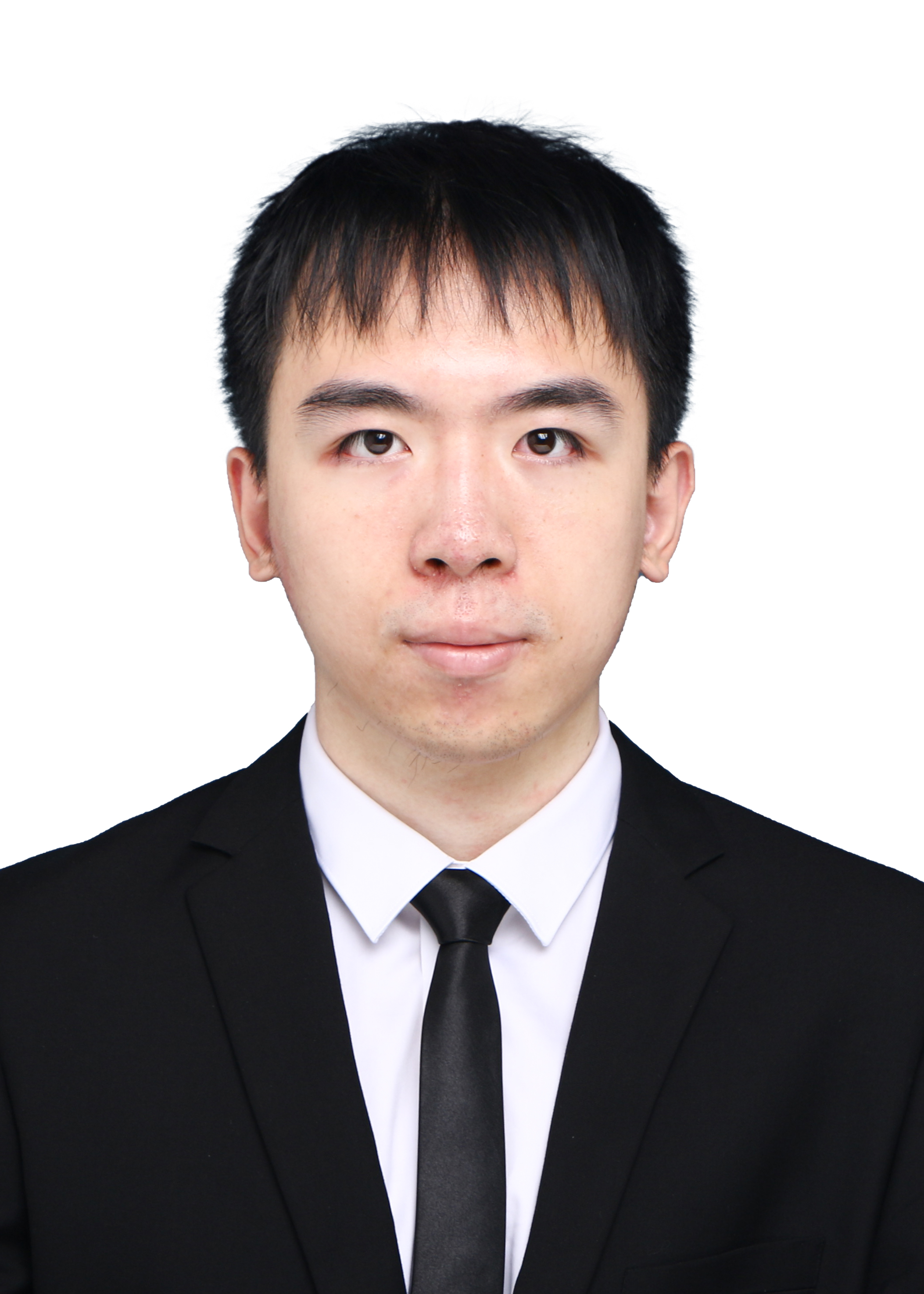}}]{Luxi Chen}
Luxi Chen received a BS degree from the Gaoling School of Artificial Intelligence, Renmin University of China, Beijing, China. He is pursuing a PhD degree in the Gaoling School of Artificial Intelligence, Renmin University of China. His research interests include deep generative models and 3D content generation.
\end{IEEEbiography}

\vspace{-5pt}

\begin{IEEEbiography}[{\includegraphics[width=1in,height=1.25in,clip,keepaspectratio]{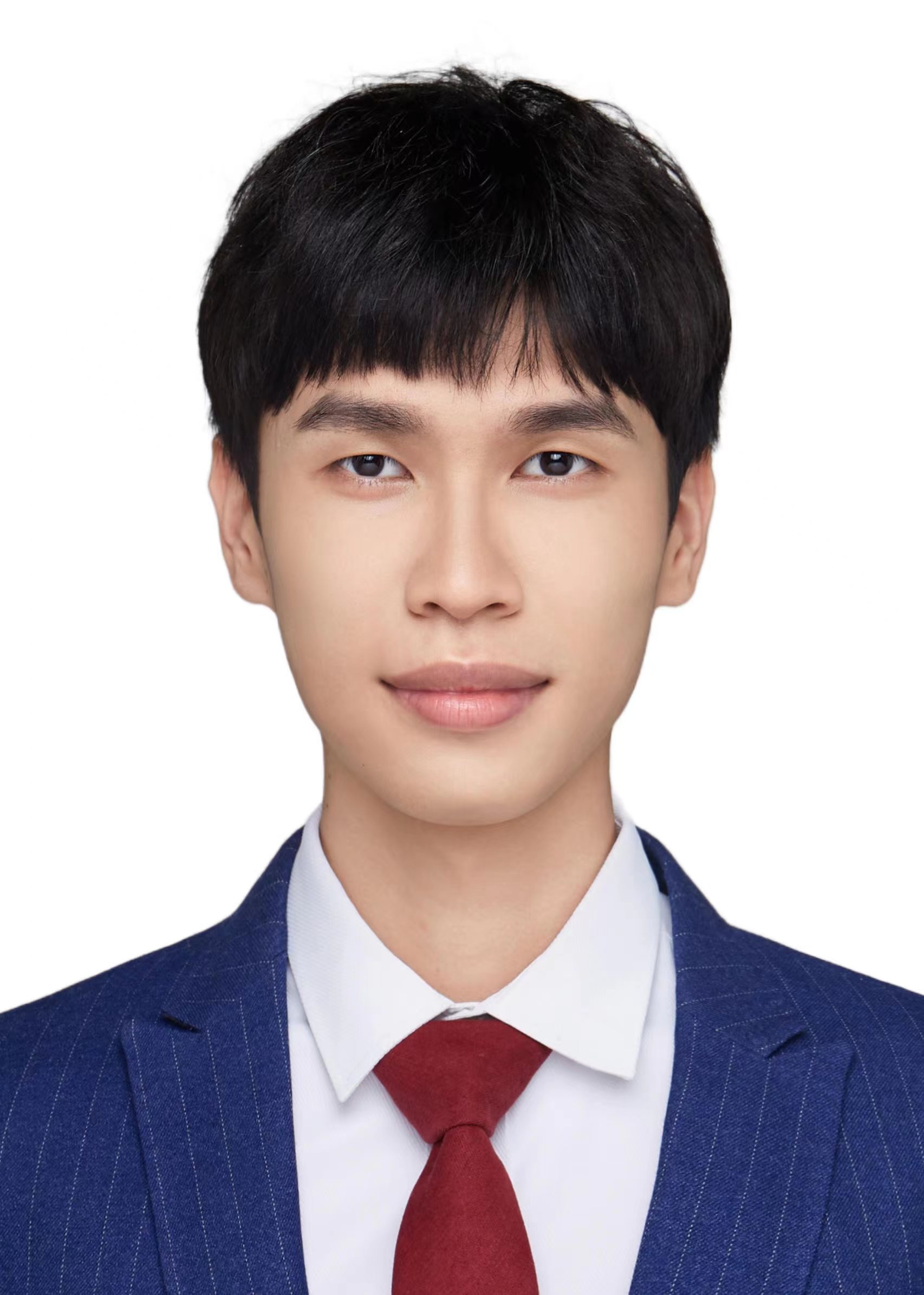}}]{Zhengyi Wang}
Zhengyi Wang received his BS degree from the Department of Computer Science and Technology, Tsinghua University, Beijing, China. He is currently working toward a PhD degree in the Department of Computer Science and Technology, at Tsinghua University, Beijing, China. His research interests include the theory and application of generative models.
\end{IEEEbiography}

\vspace{-5pt}

\begin{IEEEbiography}[{\includegraphics[width=1in,height=1.25in,clip,keepaspectratio]{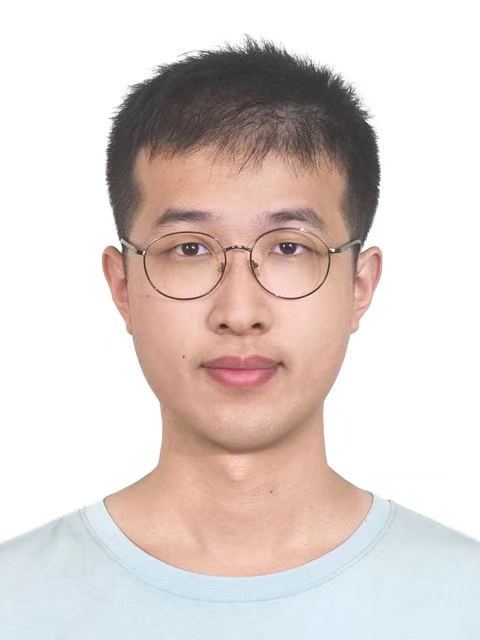}}]{Zihan Zhou}
Zihan Zhou received his BS degree from the School of Computer Science and Technology, Xidian University, Shaanxi, China. He is currently pursuing an MS
degree in the Gaoling School of Artificial Intelligence, at Renmin University of China. His research interests include 3D mesh generation and deep generative models.
\end{IEEEbiography}

\vspace{-5pt}

\begin{IEEEbiography}[{\includegraphics[width=1in,height=1.25in,clip,keepaspectratio]{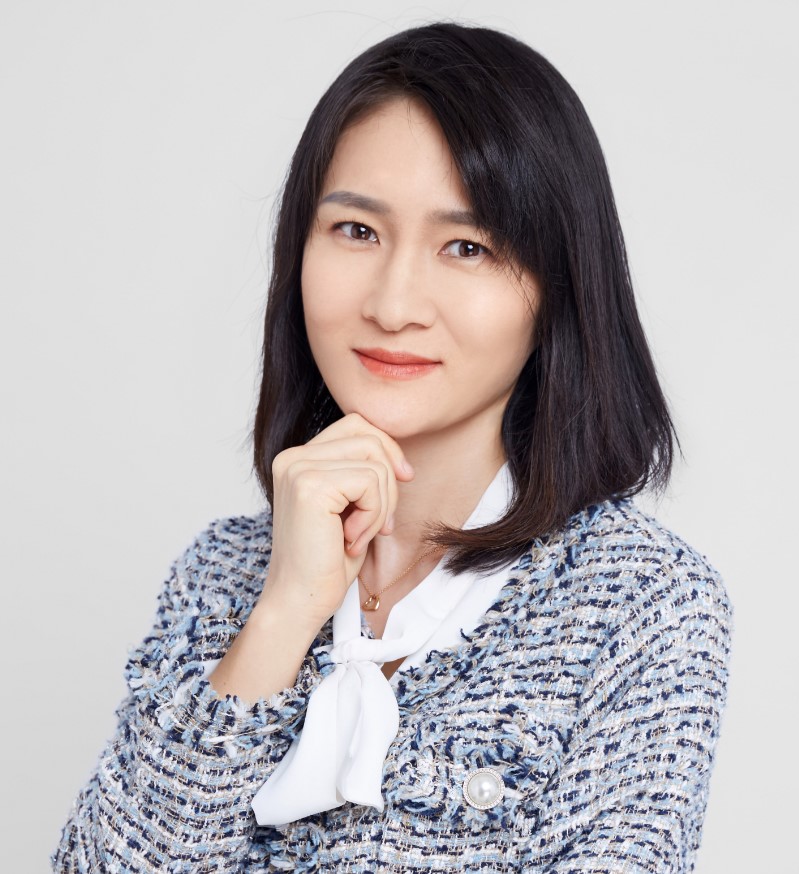}}]{Tingting Gao}
Tingting Gao is the head of the Visual Understanding and Application Center at Kuaishou. Her research interests encompass computer vision, multimodality, and the industrial application of large models. She has previously worked as a Senior Algorithm Engineer at Baidu, where she accumulated a wealth of experience in the fields of search and recommendation
\end{IEEEbiography}

\vspace{-5pt}

\begin{IEEEbiography}[{\includegraphics[width=1in,height=1.25in,clip,keepaspectratio]{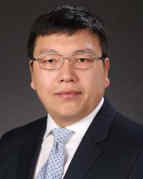}}]{Hang Su}
Hang Su (Member, IEEE) is an associate professor with the Department of Computer Science and Technology, at Tsinghua University. His research interests lie in adversarial machine learning and robust computer vision, based on which he has published more than 50 papers including CVPR, ECCV, IEEE Transactions on Medical Imaging, etc. He has served as area chair in NeurIPS and the workshop co-chair in AAAI22. He received the “Young Investigator Award” from MICCAI2012, the “Best Paper Award” in AVSS2012, and the “Platinum Best Paper Award” in ICME2018.
\end{IEEEbiography}
\vspace{-5pt}

\begin{IEEEbiography}[{\includegraphics[width=1in,height=1.25in,clip,keepaspectratio]{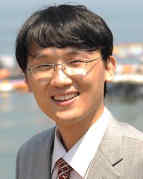}}]{Jun Zhu}
Jun Zhu (Fellow, IEEE) received the BS and PhD degrees from the Department of Computer Science and Technology, Tsinghua University, where he is currently a Bosch AI professor. He was a postdoctoral fellow and adjunct faculty with the Machine Learning Department, at Carnegie Mellon University. His research interest is primarily in developing machine learning methods to understand scientific and engineering data arising from various fields. He regularly serves as senior area chairs and area chairs at prestigious conferences, including ICML, NeurIPS, ICLR, IJCAI, and AAAI. He was selected as “AI’s 10 to Watch” by IEEE Intelligent Systems. He is a fellow of IEEE, a fellow of AAAI, and an associate editor-in-chief of IEEE Transactions on Pattern Analysis and Machine Intelligence.
\end{IEEEbiography}

\vspace{-5pt}

\begin{IEEEbiography}[{\includegraphics[width=1in,height=1.25in,clip,keepaspectratio]{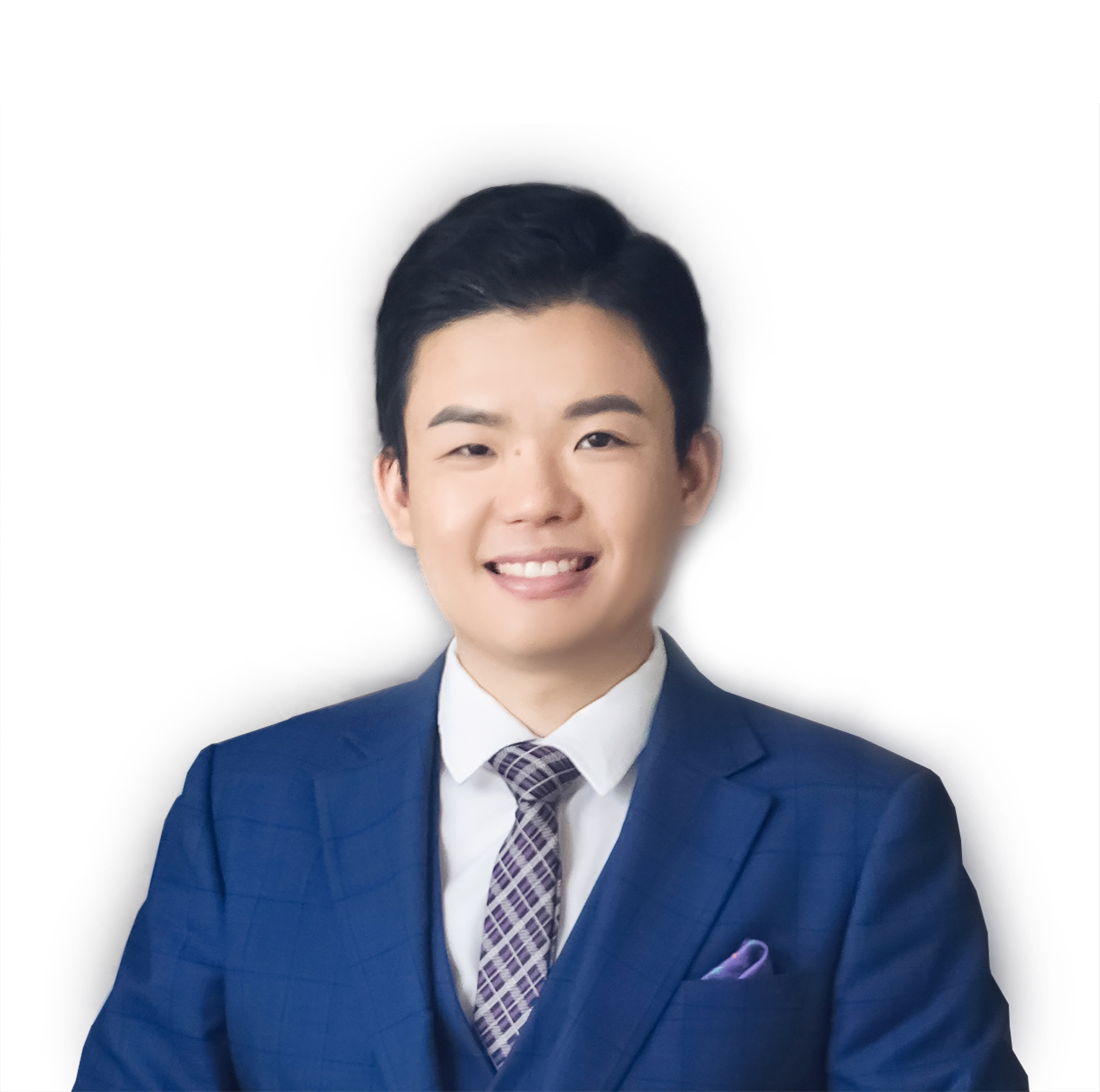}}]{Chongxuan Li}
Chongxuan Li (Member, IEEE) is an associate professor at Renmin University of China, Beijing, China. He obtained both his Bachelor's and Ph.D. degrees from Tsinghua University. His research interests include machine learning and deep generative models. His works were recognized as the Outstanding Paper Award at ICLR 2022. Moreover, he served as an associate editor for IEEE Transactions on Pattern Analysis and Machine Intelligence and area chair for NeurIPS, ICLR, and ACM MM.
\end{IEEEbiography}

\end{document}